\documentclass[lettersize,journal]{IEEEtran}

\usepackage{textcomp}
\usepackage{stfloats}
\usepackage{verbatim}
\usepackage{cite}
\usepackage[utf8]{inputenc} 
\usepackage[T1]{fontenc}    
\usepackage[colorlinks,linkcolor=blue,anchorcolor=black,citecolor=blue]{hyperref}
\usepackage{url}            
\usepackage{booktabs}       
\usepackage{amsfonts}       
\usepackage{nicefrac}       
\usepackage{microtype}      
\usepackage{graphicx}
\usepackage{subcaption}

\usepackage[table]{xcolor}
\usepackage{xcolor}         
\definecolor{Gray}{gray}{0.9}
\usepackage{threeparttable}
\usepackage{amssymb}
\usepackage{pifont}

\usepackage{amsmath}
\usepackage{amsthm}
\usepackage{array}
\usepackage{siunitx}
\usepackage{xspace}
\usepackage{longtable}
\usepackage{marvosym}
\usepackage{enumerate}
\usepackage{forest}
\usepackage{mathrsfs}
\usepackage{arydshln}
\usepackage{wrapfig}

\usepackage{makecell}
\usepackage{thmtools} 
\usepackage{thm-restate}
\usepackage{algorithm,algorithmicx,algpseudocode}
\usepackage{enumitem}
\usepackage{adjustbox}

\usepackage{multirow}

\makeatletter
\newcommand*\bigcdot{\mathpalette\bigcdot@{.5}}
\newcommand*\bigcdot@[2]{\mathbin{\vcenter{\hbox{\scalebox{#2}{$\m@th#1\bullet$}}}}}
\makeatother

\usepackage[capitalize]{cleveref}
\crefname{section}{Sec.}{Secs.}
\Crefname{section}{Section}{Sections}
\Crefname{table}{Table}{Tables}
\crefname{table}{Tab.}{Tabs.}

\newcommand\hlf[1]{\textbf{\textcolor{red}{#1}}} 
\newcommand\hls[1]{\textbf{\textcolor{blue}{#1}}}  %

\begin{document}

\title{NNG-Mix: Improving Semi-supervised Anomaly Detection with Pseudo-anomaly Generation}

\author{Hao Dong, Ga\"etan Frusque, Yue Zhao, Eleni Chatzi, and Olga Fink
\thanks{Hao Dong and Eleni Chatzi are with the Department of Civil, Environmental and Geomatic Engineering, ETH Z\"urich, 8093 Z\"urich, Switzerland (e-mail: hao.dong@ibk.baug.ethz.ch; chatzi@ibk.baug.ethz.ch).}
\thanks{Ga\"etan Frusque and Olga Fink are with the Chair of Intelligent Maintenance and Operation Systems, EPFL, 1015 Lausanne, Switzerland (e-mail: gaetan.frusque@epfl.ch; olga.fink@epfl.ch).}
\thanks{Yue Zhao is with the Thomas Lord Department of Computer Science, University of Southern California, Los Angeles, CA 90007, United States (e-mail: yzhao010@usc.edu).}}

\markboth{NNG-Mix: Improving Semi-supervised Anomaly Detection with Pseudo-anomaly Generation}%
{Shell \MakeLowercase{\textit{et al.}}: A Sample Article Using IEEEtran.cls for IEEE Journals}


\maketitle

\begin{abstract}
Anomaly detection (AD) is essential in identifying rare and often critical events in complex systems, finding applications in fields such as network intrusion detection, financial fraud detection, and fault detection in infrastructure and industrial systems. 
While AD is typically treated as an unsupervised learning task due to the high cost of label annotation, it is more practical to assume access to a small set of labeled anomaly samples from domain experts, as is the case for semi-supervised anomaly detection. Semi-supervised and supervised approaches can leverage such labeled data, resulting in improved performance. 
In this paper, rather than proposing a new semi-supervised or supervised approach for AD, we introduce a novel algorithm for generating additional pseudo-anomalies on the basis of the limited labeled anomalies and a large volume of unlabeled data. This serves as an augmentation to facilitate the detection of new anomalies. Our proposed algorithm, named Nearest Neighbor Gaussian Mixup (NNG-Mix), efficiently integrates information from both labeled and unlabeled data to generate pseudo-anomalies. We compare the performance of this novel algorithm with commonly applied augmentation techniques, such as Mixup and Cutout. We evaluate NNG-Mix by training various existing semi-supervised and supervised anomaly detection algorithms on the original training data along with the generated pseudo-anomalies. Through extensive experiments on $57$ benchmark datasets in ADBench, reflecting different data types, we demonstrate that NNG-Mix outperforms other data augmentation methods. It yields significant performance improvements compared to the baselines trained exclusively on the original training data. Notably, NNG-Mix yields up to $16.4\%$, $8.8\%$, and $8.0\%$ improvements on Classical, CV, and NLP datasets in ADBench. Our source code is available at \href{https://github.com/donghao51/NNG-Mix}{https://github.com/donghao51/NNG-Mix}.
\end{abstract}


\section{Introduction}
\IEEEPARstart{A}{nomaly} detection (AD) aims to identify irregular patterns or observations that deviate significantly from the expected behavior within a dataset. It finds applications across various domains, including social media analysis~\cite{yu2017ring}, rare disease detection~\cite{zhao2021suod}, complex systems~\cite{zhou2020siamese,frusque2023non,zhao2023dynamic,michau2021unsupervised}, and autonomous driving~\cite{bogdoll2022anomaly,SuperFusion,dong2023jras}. Due to the rarity of anomaly events and high annotation cost~\cite{han2022adbench}, AD is typically treated as an unsupervised learning problem~\cite{ruff2018deep,ergen2019unsupervised,liu2021anomaly} with the assumption that the majority of the unlabeled data in the training dataset represents normal instances. 
Unsupervised AD algorithms typically learn a model or distribution that accurately describes normal behavior and detect deviations from this description as anomalies. 
Various unsupervised methods have been proposed, including shallow machine learning approaches like One-Class SVM~\cite{scholkopf2001estimating}, Kernel Density Estimation~\cite{vandermeulen2013consistency}, Isolation Forest~\cite{liu2008isolation}, and Empirical Cumulative Distribution Functions~\cite{li2022ecod} along with deep-learning-based approaches such as Deep Support Vector Data Description (DeepSVDD)~\cite{ruff2018deep} and Deep Autoencoding Gaussian Mixture Model (DAGMM)~\cite{zong2018deep}. 
Recent literature indicates that unsupervised methods are effective only when their assumptions align with the characteristics of the considered dataset. For instance, the $k$ nearest neighbor~\cite{ramaswamy2000efficient} functions optimally as a global detector when the dataset indeed contains global anomalies, and local outlier factor \cite{breunig2000lof} performs well with local outliers only.
However, this scenario is not consistently mirrored in real-world applications. The real outliers can be a mix of multiple types of outliers and are difficult to identify beforehand.

Beyond the option of adopting fully unsupervised methods, accessing some labeled anomalies in addition to unlabeled data is feasible in many real-world applications, such as the inspection of complex systems~\cite{frusque2023non,zhao2023dynamic,michau2021unsupervised}.
A recent study~\cite{han2022adbench} demonstrated that the introduction of a small number of anomalies, whether through a semi-supervised or supervised approach, consistently resulted in improvements in anomaly detection with respect to the unsupervised case.
Anomalies are often associated with safety-critical events, leading to their rare occurrences. Therefore, the majority of samples in a dataset usually represent normal data. While some anomalous samples may have been identified and labeled by domain experts, it is important to note that not all anomalous samples may have been recognized or captured.
The number of labeled anomalies is often scarce compared to the unlabeled data. With only a small number of such labeled samples available, several semi-supervised AD algorithms have been proposed, including Deep Semi-supervised Anomaly Detection (DeepSAD)~\cite{DeepSAD}, Pairwise Relation prediction Network (PReNet)~\cite{pang2023deep}, Deviation Networks (DevNet)~\cite{pang2019deep}, MPAD~\cite{mpad}, FEAWAD~\cite{zhou2021feature}, and Extreme Gradient Boosting Outlier Detection (XGBOD)~\cite{zhao2018xgbod}. For instance, DeepSAD~\cite{DeepSAD} introduces a novel loss function to bring normal data closer to a fixed center and position anomalies farther away. In contrast, supervised methods such as MLP~\cite{rosenblatt1958perceptron}, Feature Tokenizer + Transformer (FTTransformer)~\cite{gorishniy2021revisiting}, and Categorical Boosting (CatB)~\cite{prokhorenkova2018catboost} treat anomaly detection as binary classification. However, they are limited to detecting known types of anomalies~\cite{aggarwal2017introduction}, as the training dataset may not cover all possible types of anomalies. Moreover, for both semi-supervised and supervised approaches, deep learning models might struggle to learn effective representations if labeled anomalies are extremely limited.

Unlike existing solutions that primarily focus on creating novel semi-supervised~\cite{DeepSAD,pang2023deep,zong2018deep} or supervised~\cite{rosenblatt1958perceptron,gorishniy2021revisiting,prokhorenkova2018catboost,massoli2021mocca} algorithms to better leverage the use of limited labeled data for anomaly detection, our approach focuses on harnessing the potential of generating pseudo-anomalies to improve AD performance. 
More specifically, we generate pseudo-anomalies by leveraging the limited set of labeled anomalies and a large amount of unlabeled data. Then we train various semi-supervised or supervised AD algorithms on the original training data in conjunction with the generated pseudo-anomalies. With the availability of more anomaly data for training, we demonstrate performance improvements compared with the original AD algorithms.

\begin{figure*}[t!]
  \centering  \includegraphics[width=0.7\linewidth]{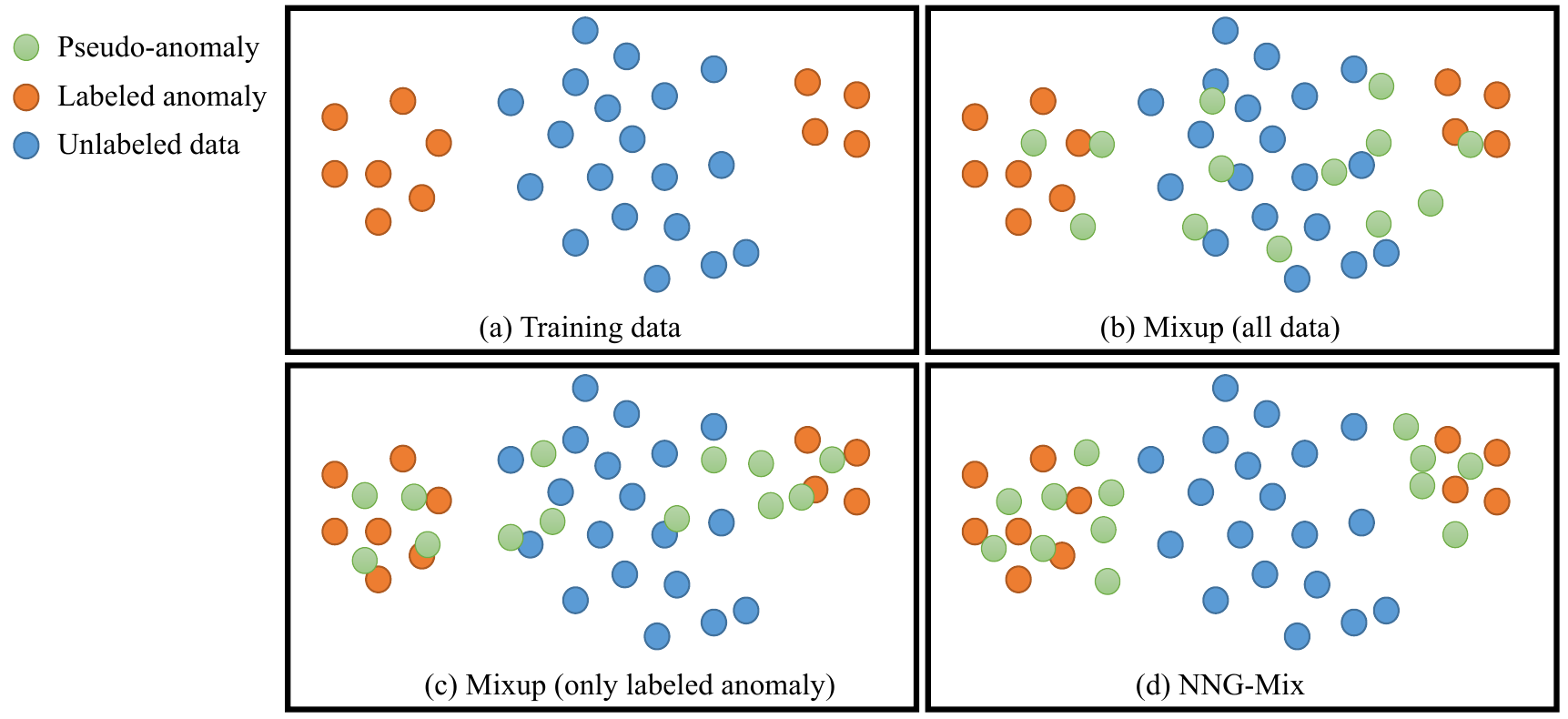}
   \caption{Pseudo-anomaly generation using Mixup~\cite{zhang2018mixup} and our proposed NNG-Mix. Mixup using all training data introduces noise samples within the distribution of unlabeled data, while using only labeled anomalies underestimates the information from unlabeled data and also injects some noise samples that are within unlabeled data. In contrast, our NNG-Mix fully exploits the information from both labeled anomalies and unlabeled data samples to generate pseudo-anomalies effectively. }
   \label{fig:data-gen}
\end{figure*}

Pseudo-anomaly generation is similar to data augmentation, a technique employed to mitigate overfitting by training models on slightly modified copies of existing data. Well-established data augmentation methods like Mixup~\cite{zhang2018mixup}, Cutout~\cite{devries2017cutout}, CutMix~\cite{yun2019cutmix}, and the addition of random Gaussian noises can be readily adapted for pseudo-anomaly generation. For example, Mixup creates new samples by the convex combinations of sample pairs.
However, traditional data augmentation methods commonly adopted in computer vision may not be optimal for the anomaly detection task. If we mix the data randomly sampled from all training data using Mixup, there may exist scenarios where generated pseudo-anomalies fall within the distribution of unlabeled data. \cref{fig:data-gen} (a) shows normal samples surrounded by two different anomaly types. Mixup generates new anomaly samples that overlap with the distribution of the unlabeled samples, as demonstrated in~\cref{fig:data-gen} (b). 
Given the general assumption that the majority of unlabeled data is normal, the introduction of such generated pseudo-anomalies could significantly hinder the performance of anomaly detection.
Conversely, if we only mix the data randomly sampled from labeled anomalies, as illustrated in~\cref{fig:data-gen} (c), the output space for generated pseudo-anomalies becomes excessively constrained, because of the limited utilization of information from unlabeled data. Mixing only the labeled anomalies can also introduce noise samples that are within the distribution of unlabeled data, potentially adversely impacting the performance of anomaly detection.

In this paper, we propose a \textbf{N}earest \textbf{N}eighbor \textbf{G}aussian \textbf{Mix}up (NNG-Mix) algorithm for generating pseudo-anomalies that optimally exploit information from both labeled anomalies and unlabeled data. To prevent the generation of anomaly samples that are within the distribution of the normal data, we exclusively mix the anomalies with their top $k$ nearest neighbors from the unlabeled data. Additionally, we introduce random Gaussian noise to the samples before applying Mixup, thereby expanding the potential space for generating pseudo-anomalies. As shown in~\cref{fig:data-gen} (d), NNG-Mix effectively generates pseudo-anomalies without introducing undesired noise samples. 
Our main contributions are the following:
\begin{itemize}[leftmargin=*]
\setlength\itemsep{0em}
\item \textbf{Novel Anomaly Detection Framework}. We investigate improving semi-supervised anomaly detection performance from a novel viewpoint, by generating additional pseudo-anomalies based on the limited labeled anomalies and a large amount of unlabeled data. 
\item \textbf{In-depth Analysis}. We systematically evaluate four different data augmentation methods for pseudo-anomaly generation.
\item \textbf{Novel Algorithm}. We introduce NNG-Mix, a simple and effective pseudo-anomaly generation algorithm, that optimally utilizes information from both labeled anomalies and unlabeled data.
\item \textbf{Evaluation and Effectiveness}. We conduct extensive experiments on $57$ datasets in ADBench, reflecting diversified domains, including applications in healthcare, audio and language processing, 
finance, etc., and on five semi-supervised and supervised anomaly detection algorithms. Our NNG-Mix yields substantial performance improvements compared to different baseline data augmentation methods.
\end{itemize}

\section{Related Work}
\noindent\textbf{Anomaly Detection} aims to identify data instances that deviate significantly from the majority of data objects. Different types of AD methods have been proposed depending on the availability of labels, including unsupervised, semi-supervised, and supervised methods. Unsupervised AD methods usually assume no access to labeled data and often suggest that anomalies are situated in low-density regions~\cite{bergman2019classification,zhao2019pyod}. For example, Local Outlier Factor (LOF)~\cite{breunig2000lof} measures the local deviation of the density of a given sample with respect to its neighbors. DeepSVDD~\cite{ruff2018deep} trains a neural network that minimizes the volume of a hypersphere to encapsulate the network representations of the data, driving the network to extract the common factors of variation. In a recent study by ~\cite{han2022adbench}, various unsupervised anomaly detection approaches were compared across multiple datasets. The findings revealed that no single algorithm consistently outperformed others across all datasets. Furthermore, the study demonstrates that the introduction of a small number of anomalies, whether through a supervised or semi-supervised approach, resulted in consistent improvements in the AD tasks.

Supervised AD methods consider anomaly detection as binary classification, and prevailing classifiers such as MLP~\cite{rosenblatt1958perceptron}, FTTransformer~\cite{gorishniy2021revisiting}, and CatB~\cite{prokhorenkova2018catboost} are frequently used for this purpose. MLP~\cite{rosenblatt1958perceptron} uses the binary cross entropy loss to update network parameters, while FTTransformer~\cite{gorishniy2021revisiting} presents an effective adaptation of the Transformer architecture~\cite{vaswani2017attention} specifically designed for tabular data. CatB~\cite{prokhorenkova2018catboost} operates as a fast, scalable, high-performance gradient boosting technique based on decision trees. One of the main challenges encountered when using supervised algorithms for anomaly detection is the limited availability of anomalous data, often failing to encompass the entire range of potential anomalies.

Hence, semi-supervised methods~\cite{DeepSAD,zhao2018xgbod} stand out as an effective strategy, utilizing partial supervision from labeled data to enhance anomaly detection performance. There are different types of directions for semi-supervised approaches. The Extreme Gradient Boosting Outlier Detection (XGBOD)~\cite{zhao2018xgbod}, for example, augments the feature space by extracting useful representations using multiple unsupervised AD algorithms and concatenating them together as features to enhance predictive capabilities. DeepSAD~\cite{DeepSAD} integrates a term for the labeled anomalies in the loss function to leverage the supervision from labeled data. FEAWAD~\cite{zhou2021feature} enhances AD performance by utilizing hidden representations, reconstruction residual vectors, and reconstruction errors from an autoencoder as the new representations. They use labeled anomalies to force the anomaly scores of normal samples to be close to $0$ while the anomaly scores for labeled anomalies are away from $0$ by a predefined margin. Different from previous methods, RoSAS~\cite{xu2023rosas} proposes a mass interpolation method to create samples labeled with continuous abnormal degrees, complemented by a feature learning-based objective that serves as an optimization constraint to regularize the network.

In this study, our objective is to enhance the anomaly detection performance by generating additional pseudo-anomalies as training data to improve the generalization ability of different AD methods. This approach is designed to be flexible and can be integrated with any supervised or semi-supervised AD method. To demonstrate this flexibility and versatility, and assess the performance of the proposed anomaly generation framework, we have combined our augmentation algorithm with current State-of-the-Art (SOTA) semi-supervised and supervised AD methods  in various experiments. 


\noindent\textbf{Data Augmentation} is a technique in deep learning to mitigate overfitting by generating modified versions of existing data samples to increase diversity and improve generalization and has been widely used in many areas like computer vision (CV)~\cite{yang2022image} and natural language processing (NLP)~\cite{feng2021survey}. In CV, basic data augmentation methods~\cite{chen2020simple} involve image manipulations, including cropping, resizing, color distortion, rotation, addition of Gaussian noise and blur, and techniques like Cutout~\cite{devries2017cutout}. Another method involves different mixing strategies, which combine two or more images or their sub-regions. For example, Mixup~\cite{zhang2018mixup} and its variants~\cite{yao2022cmix,li2022who} perform convex combinations between image pairs and their labels and encourage the model to behave linearly between training samples. This behavior can lead to a smoother and more robust model that generalizes better to unseen data. CutMix~\cite{yun2019cutmix} replaces removed regions with a patch from another image. Advanced approaches such as AutoAugment~\cite{cubuk2019autoaugment} automatically search for improved data augmentation policies. The Perturbation Learning Based Anomaly Detection approach (PLAD)~\cite{cai2022perturbation} utilizes  small perturbations to alter  normal data. The method then  applies a classifier to differentiate between  the original normal data and the perturbed data, effectively categorizing them  into two distinct  classes. In NLP, basic augmentation methods involve random insertion, deletion, and swap~\cite{wei2019eda}, while mixed sample data augmentation explores interpolating inner components~\cite{guo2020nonlinear}. Besides CV and NLP, the Calibrated One-class classification-based Unsupervised Time
series Anomaly detection (COUTA)~\cite{xu2024calibrated} generates anomalous samples through  perturbation to mimic  abnormal behaviors in time series data. Similarly, GOAD~\cite{bergman2020classification} employs affine transformations to shift the data into different subspaces and learns a feature space where the separation between classes is greater than the separation within classes.
While many data augmentation methods can be adapted for pseudo-anomaly generation, their optimality might be limited without considering the specific characteristics of anomaly detection tasks. In this work, we propose a novel pseudo-anomaly generation algorithm that fully leverages information from both labeled anomalies and unlabeled data.

\noindent\textbf{Self-supervised AD} is motivated by the recent success of self-supervised learning~\cite{jing2020self}, where a model learns a generalizable representation from unlabelled data by solving one or several pretext tasks. These tasks use data as labels and are not directly related to the final downstream task. In self-supervised AD, the pretext task is designed to guide the model in learning a representation tailored for anomaly detection. For example, CutPaste~\cite{li2021cutpaste} generates local irregular patterns in the training phase and detects similar irregularities, representative of unseen real defects, during the testing phase. In a different line of research, SSL-OE~\cite{hendrycks2019using} uses an auxiliary rotation loss to improve the robustness of AD models. SSD~\cite{sehwag2021ssd} uses contrastive learning~\cite{winkens2020contrastive} for unsupervised representation learning adopting by a Mahalanobis distance in the feature space as the probabilistic metric that enables anomaly detection. While these approaches have shown promising results in AD, they are entirely unsupervised and only effective when their assumptions align with the characteristics of the considered dataset. Contrary to these approaches, our setting has access to a small set of labeled anomalies, making it more realistic.


\section{Methodology}
\subsection{Preliminaries and Problem Definition}
In this work, we aim to generate additional pseudo-anomalies based on the limited labeled anomalies and a large volume of unlabeled data as an augmentation to facilitate the detection of new anomalies. Let us consider a scenario where we have a limited number of labeled anomaly data $\mathcal{A}$ and a large amount of unlabeled data $\mathcal{H}$, where the majority of data samples in $\mathcal{H}$ represents normal data and we treat them as normal data during training. The label of sample is denoted by $y$, where $y=1$ represents anomalies and $y=0$ represents normal data. The primary objective of semi-supervised and supervised AD methods is to train a model $\mathbf{F}$ on the combined dataset $\{\mathcal{A}, \mathcal{H}\}$ to generate an anomaly score for each sample. This work delves into generating additional pseudo-anomaly data $\mathcal{D}$ using available data $\mathcal{A}$ and $\mathcal{H}$. Subsequently, $\mathbf{F}$ is trained on the extended dataset $\{\mathcal{A}, \mathcal{H}, \mathcal{D}\}$ with the aim of enhancing anomaly detection performance and improving generalization.


\subsection{Baseline Pseudo-anomaly Generation Methods}
We evaluate the performance improvement and generalization ability of other commonly applied data augmentation approaches that are not specifically tailored for anomaly generation, including Mixup~\cite{zhang2018mixup}, Cutout~\cite{devries2017cutout}, CutMix~\cite{yun2019cutmix}, and random Gaussian noises. Then, we introduce the proposed novel Nearest Neighbor Gaussian Mixup method to enhance the effectiveness of pseudo-anomaly generation.

\noindent\textbf{Mixup for Pseudo-anomaly Generation. }
Mixup~\cite{zhang2018mixup} generates a new sample $(\mathbf{d}, y)$ by creating convex combinations of a pair of samples and their respective labels:
\begin{equation}
\begin{split}
  \mathbf{d} &= \lambda \mathbf{a_1} + (1 - \lambda) \mathbf{a_2}, \\
 y &= \lambda y_1 + (1 - \lambda) y_2, \\
\end{split}
\label{eq:mixup}
\end{equation}
where $(\mathbf{a_1}, y_1)$ and $(\mathbf{a_2}, y_2)$ are two samples randomly drawn from the
training data, and $\lambda \sim \text{Beta}(\alpha, \alpha)$, for $\alpha \in (0, \infty)$. 
In our experiments, $(\mathbf{a_1}, y_1)$ and $(\mathbf{a_2}, y_2)$ are exclusively drawn from the labeled anomaly data $\mathcal{A}$. We show in our ablation study that applying Mixup between anomaly and unlabeled data without any constraints yields inferior performance. We derive pseudo-anomaly $\mathbf{d}$ using $\mathbf{d} = \lambda \mathbf{a_1} + (1 - \lambda) \mathbf{a_2}$. Concerning the label $y$, instead of utilizing the convex combinations of $1$ and $0$, we assume $y$ to be $1$ for all samples, which implies that all generated data are assumed to be anomalies.

\noindent\textbf{Cutout for Pseudo-anomaly Generation. }
Cutout~\cite{devries2017cutout} randomly masks out some square or continuous regions in a sample. Given $\mathbf{a_1}$ drawn from $\mathcal{A}$, we generate the pseudo-anomaly $\mathbf{d}$ as follows: 
\begin{equation}
\begin{split}
  \mathbf{d} & = \mathbf{m} \odot \mathbf{a_1}, \\
\end{split}
\label{eq:cutout}
\end{equation}
where $\mathbf{m} \in \{0,1\}^{m}$ denotes a binary mask indicating where to drop out from $\mathbf{a_1}$, and $\odot$ represents element-wise multiplication, $m$ is the length of $\mathbf{a_1}$.

\noindent\textbf{CutMix for Pseudo-anomaly Generation. }
CutMix~\cite{yun2019cutmix} involves cutting and pasting patches among training samples, where the ground truth labels are also mixed proportionally based on the area of the patches. This process is mathematically represented as:
\begin{equation}
\begin{split}
  \mathbf{d} & = \mathbf{m} \odot \mathbf{a_1} + (\mathbf{1}- \mathbf{m}) \odot \mathbf{a_2} \\
  y & = \lambda y_1 + (1-\lambda) y_2, \\
\end{split}
\label{eq:cutmix}
\end{equation}
where $\mathbf{m} \in \{0,1\}^{m}$ denotes a binary mask indicating areas to be replaced and filled from two samples, and $\mathbf{1}$ represents a binary mask filled with ones. 
Similar to Mixup, in our experiment, the pairs $(\mathbf{a_1}, y_1)$ and $(\mathbf{a_2}, y_2)$ are drawn exclusively from $\mathcal{A}$. We generate pseudo-anomaly samples using $\mathbf{d} = \mathbf{m} \odot \mathbf{a_1} + (\mathbf{1}- \mathbf{m}) \odot \mathbf{a_2}$ while fixing $y$ to be 1.

\noindent\textbf{Gaussian Noise for Pseudo-anomaly Generation. }
In this baseline method, we generate pseudo-anomaly $\mathbf{d}$ by adding random Gaussian noises (GN) to the labeled anomaly data $\mathbf{a_1}$ from $\mathcal{A}$ as represented by the equation:
\begin{equation}
\begin{split}
  \mathbf{d} = \mathbf{a_1} + \boldsymbol{\epsilon}, \\
\end{split}
\label{eq:gaussian}
\end{equation}
where $\boldsymbol{\epsilon} $ is drawn from a centered Gaussian distribution with a standard deviation of $\boldsymbol{\sigma}$, denoted as $\boldsymbol{\epsilon} \sim \mathcal{N}(\mathbf{0}, \boldsymbol{\sigma})$.

\subsection{Nearest Neighbor Gaussian Mixup}
\textbf{Motivation}. While Mixup offers the potential to generate new samples by blending unlabeled and anomaly samples, this approach poses the risk of producing mislabeled samples. For instance, C-Mixup~\cite{yao2022cmix} demonstrates that directly applying vanilla Mixup on regression tasks can lead to arbitrarily incorrect labels and proposes an adjustment in the sampling probability based on label similarity.
In our scenario, as illustrated in Fig.~\ref{fig:data-gen} (b) and (c), vanilla Mixup introduces a considerable number of noisy samples within the distribution of the unlabeled data. This influx of noise could significantly hinder anomaly detection performance. 
Moreover, the exclusive mixing of labeled anomaly samples presents the drawback of failing to leverage information from the unlabeled data, which could be valuable in producing more relevant anomalies. To effectively utilize the unlabeled data and avoid generating unwanted noise samples, we propose Nearest Neighbor Gaussian Mixup, a simple yet effective pseudo-anomaly generation algorithm.

\begin{figure*}[t!]
  \centering  \includegraphics[width=0.6\linewidth]{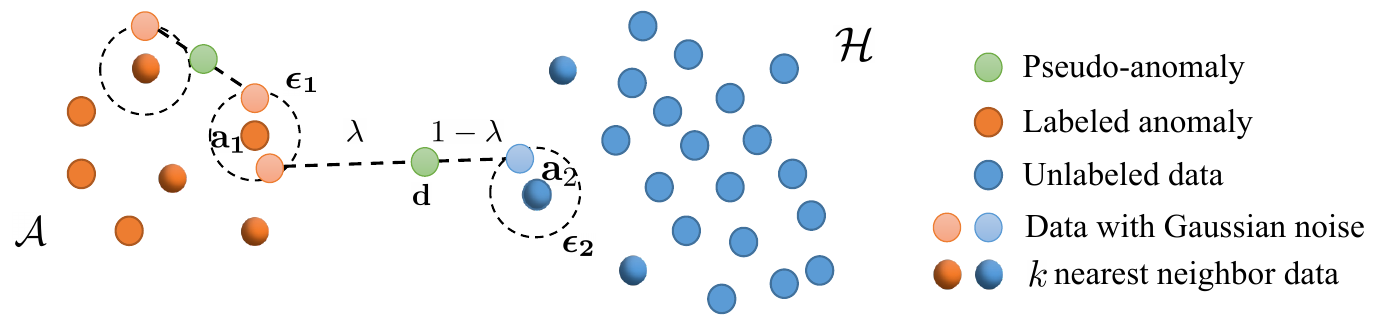}
   \caption{Nearest Neighbor Gaussian Mixup makes good use of information from both labeled anomalies and unlabeled data to generate pseudo-anomalies effectively.}
   \label{fig:illus}
\end{figure*}

\textbf{Proposed Approach}. Instead of randomly selecting an anomaly sample from $\mathcal{A}$ and an unlabeled sample from $\mathcal{H}$ to mix, our approach involves obtaining a labeled anomaly sample $\mathbf{a_1}$ from $\mathcal{A}$ and identifying its top $k$ nearest neighbors, denoted as $\mathcal{M}$ from the unlabeled set $\mathcal{H}$. 
We then randomly select one unlabeled sample $\mathbf{a_2}$ from $\mathcal{M}$ for mixup with $\mathbf{a_1}$. 
This process aims to prevent the generation of anomalies that reside within the distribution of unlabeled data, as shown in Fig.~\ref{fig:data-gen} (d). 
Our algorithm also has a $50\%$ probability of choosing the top $k$ nearest neighbors from $\mathcal{A}$. This ensures the combination of two anomalies belonging to the same or similar anomaly category, thereby alleviating the generation of unrealistic artificial anomalies that lie within the distribution of unlabeled data. Before applying the convex combination of the two samples using the parameter $\lambda$ drawn from a beta distribution, we augment both $\mathbf{a_1}$ and $\mathbf{a_2}$ by adding random Gaussian noises $\boldsymbol{\epsilon_1}$ and $\boldsymbol{\epsilon_2}$  with a zero mean and a standard deviation of $\boldsymbol{\sigma}$. This augmentation broadens the potential space for generating pseudo-anomalies. 
Then, NNG-Mix can be defined by the following equations:
\begin{align}
&\mathbf{d} = \lambda (\mathbf{a}_1 + \boldsymbol{\epsilon_1}) + (1-\lambda)(\mathbf{a}_2 + \boldsymbol{\epsilon_2}) \\
& \mathbf{a}_2 \sim {\rm kNN}(\mathbf{a}_1 , \mathcal{A}, k) \hspace{0.5cm} \text{if mixed with labeled anomaly} \nonumber \\
& \mathbf{a}_2 \sim {\rm kNN}(\mathbf{a}_1 , \mathcal{H}, k) \hspace{0.5cm} \text{if mixed with unlabeled data} \nonumber \\
 &\boldsymbol{\epsilon_1},\boldsymbol{\epsilon_2} \sim \mathcal{N}(\mathbf{0}, \boldsymbol{\sigma}) \nonumber 
\end{align}
Here, $ {\rm kNN}(\mathbf{a}_1 , \mathcal{A}, k) $ and $ {\rm kNN}(\mathbf{a}_1 , \mathcal{H}, k) $ refer to the selection of one of the $k$ nearest neighbors of the sample $\mathbf{a}_1 $ from the anomaly data $\mathcal{A}$ and unlabeled data $\mathcal{H}$, respectively. The choice of a nearest neighbor from the $k$ is made randomly. 
An illustration of NNG-Mix is provided in~\cref{fig:illus}, and the pseudo-code for generating relevant anomalies using NNG-Mix is presented in~\cref{alg:NNG-Mix}. We use a Uniform distribution with a threshold of $0.5$ to ensure  a $50\%$ probability of selecting  the top $k$ nearest neighbors from either $\mathcal{A}$ or $\mathcal{H}$. 


\begin{figure}[ht]
  \centering
\begin{minipage}[t]{0.45\textwidth}
\begin{algorithm}[H]
  \caption{Nearest Neighbor Gaussian Mixup} \label{alg:NNG-Mix}
  \small
  \textbf{Input: }Anomaly data $\mathcal{A}$, unlabeled data $\mathcal{H}$, generated pseudo-anomaly $\mathcal{D}$, hyper-parameters $\alpha$, $k$ and $\boldsymbol{\sigma}$. 
  \begin{algorithmic}[1]
    \vspace{.04in}
    \State $\mathcal{D} = \{\}$ 
    \For{$i=1, \dotsc, N$}
      \State Sample anomaly $\mathbf{a_1}$ from $\mathcal{A}$
      \If {Uniform(0, 1) $>$ 0.5}
      \State $\mathcal{M}$ = {\rm kNN}($\mathbf{a_1}, \mathcal{A}, k$)
      \Else
      \State $\mathcal{M}$ = {\rm kNN}($\mathbf{a_1}, \mathcal{H}, k$)
      
      \EndIf
      \State Sample $\mathbf{a_2}$ from $\mathcal{M}$
      
      \State $\boldsymbol{\epsilon_1},\boldsymbol{\epsilon_2} \sim \mathcal{N}(\mathbf{0}, \boldsymbol{\sigma})$
      \State $\mathbf{a_1} = \mathbf{a_1} + \boldsymbol{\epsilon_1}$
      \State $\mathbf{a_2} = \mathbf{a_2} + \boldsymbol{\epsilon_2}$
      \State $\lambda$ = Beta($\alpha$, $\alpha$)
      \State $\mathbf{d} = \lambda  \mathbf{a_1} + (1-\lambda)  \mathbf{a_2}$
      \State $\mathcal{D} = \{ \mathcal{D} ; \mathbf{d}\}$
    \EndFor
    \State \textbf{return} $\mathcal{D}$
    \vspace{.04in}
  \end{algorithmic}
\end{algorithm}
\end{minipage}
\end{figure}

\section{Experiments}
We conducted experiments on ADBench~\cite{han2022adbench}, a comprehensive tabular anomaly detection benchmark. This benchmark comprises $47$ widely-used real-world Classical datasets and an additional $10$ more complex datasets from computer vision (CV) and natural language processing (NLP) domains. 
These domains cover a broad spectrum of diverse applications such as healthcare (e.g., disease diagnosis), audio and language processing (e.g., speech recognition), image processing (e.g., object identification), finance (e.g., financial fraud detection), and more. Given the diverse range of applications, ADBench serves as an ideal benchmark for systematically evaluating the performance of various pseudo-anomaly generation algorithms.

\subsection{Experimental Setup}
We aim to enhance the performance of anomaly detection algorithms in scenarios with limited availability of labeled anomalies. In particular, we evaluate both semi-supervised and supervised anomaly detection algorithms under this setup.  In our experiments, we consider scenarios containing only $10\%$, $5\%$, or even $1\%$ of labeled anomalies. To mitigate this scarcity, we employ various pseudo-anomaly generation algorithms, including Mixup, Cutout, CutMix, Gaussian Noise, and our proposed NNG-Mix. In addition, Non-parametric Outlier Synthesis (NPOS)~\cite{tao2023nonparametric}, and RoSAS~\cite{xu2023rosas} are selected as comparison methods for our study. NPOS generates artificial out-of-distribution training data without assuming  any specific distribution for  the in-distribution embeddings. RoSAS uses a strategy similar  to Mixup, using interpolation between labeled anomalies and unlabeled data to enhance learning. By applying different augmentation algorithms, we generate $M \times$ more pseudo-anomalies than the original labeled anomalies, significantly increasing the number of anomalies in the training dataset.  Subsequently, we incorporate these generated pseudo-anomalies into the original training data for training the semi-supervised and supervised anomaly detection algorithms. The expectation is that performance will be improved with the availability of a larger volume of anomaly data for training. 

For our experiments, we choose SOTA semi-supervised algorithms DeepSAD~\cite{DeepSAD} and XGBOD~\cite{zhao2018xgbod}, and supervised algorithms MLP~\cite{rosenblatt1958perceptron}, FTTransformer~\cite{gorishniy2021revisiting}, and CatB~\cite{prokhorenkova2018catboost}, based on their reported performances on ADBench~\cite{han2022adbench}. All algorithms are used with their default hyperparameter settings, as reported in the original papers for fair comparison. 

We evaluate different algorithms separately on the Classical, CV, and NLP datasets within ADBench to demonstrate the efficacy of pseudo-anomaly generation across diverse data types. The training set comprises $70\%$ of data, while the remaining $30\%$ is allocated for testing. Evaluation is conducted using the Area Under Receiver Operating Characteristic Curve (AUCROC) as the standardized metric for all experiments. 

For the pseudo-anomaly generation methods, we set the parameters as follows: Mixup~\cite{zhang2018mixup}: $\lambda \sim \text{Beta}(\alpha, \alpha)$ with $\alpha=0.2$. Cutout~\cite{devries2017cutout}: Dropout ratio is sampled between $0.1$ and $0.3$, with a randomly selected starting point for the dropout. CutMix~\cite{yun2019cutmix}: Mask ratio is sampled between $0.1$ and $0.3$, with a randomly selected starting point for the cutting and mixing process. Random Gaussian noise: $\boldsymbol{\sigma}=0.01$. RoSAS~\cite{xu2023rosas}: $\lambda \sim \text{Beta}(\alpha, \alpha)$ with $\alpha=0.5$. For NPOS~\cite{tao2023nonparametric}, we use its default parameter setup. Our proposed NNG-Mix: $\alpha=0.2$, $k=10$, and $\boldsymbol{\sigma}=0.01$.  Additionally, we perform an ablation study to analyze the sensitivity of the parameters.

\subsection{Results}

\begin{table*}[t!]
\centering
\resizebox{0.7\linewidth}{!}{
\begin{tabular}{|c|c|c|c|c|c|c|c|c|c|c|c|c|c|c|c|c|c|}
\hline \multirow{2}{*}{ Algorithm } & Baseline & \multicolumn{1}{c|}{ NPOS }& \multicolumn{1}{c|}{ RoSAS } &\multicolumn{1}{c|}{ Mixup } & \multicolumn{1}{c|}{ Cutout } & \multicolumn{1}{c|}{ CutMix } & \multicolumn{1}{c|}{ GN } & \multicolumn{1}{c|}{ NNG-Mix } \\
\cline { 2 - 17 } & $1 \%$  & $10 \times$& $10 \times$  & $10 \times$  & $10 \times$  & $10 \times$  & $10 \times$  & $10 \times$\\
\hline DeepSAD & 0.766 & 0.764  & 0.772&\hls{0.786}  &0.780 &0.782  &0.782  &\hlf{0.791} \\ 
\hline XGBOD & 0.815  & 0.804 & 0.788&0.785  &0.798  &0.816 &\hls{0.837} &\hlf{0.838} \\
\hline MLP &  0.600& 0.640& 0.724 & 0.758&0.748  &\hls{0.763} &0.751  &\hlf{0.764} \\  
\hline FTTransformer & 0.719  & 0.753&0.735 & \hls{0.781}  &0.757  &0.778 & 0.763 &\hlf{0.783} \\ 
\hline CatB &  0.851&0.835 & 0.814& 0.846& 0.836&0.844 &\hls{0.859}&\hlf{0.861} \\
\hline \textit{Average} & 0.750&  0.759&0.767 &0.791&0.783& 0.797& \hls{0.798}&\hlf{0.807}   \\
\hline
\end{tabular}
}
\caption{Results on Classical datasets with $1\%$ available labeled anomalies and $10 \times$ pseudo-anomaly generation. The best results are in \hlf{red} and the second best results are in \hls{blue}.}
\label{tab:cls-0.01-10-new}
\end{table*}
\begin{table*}[ht!]
\centering
\resizebox{0.7\linewidth}{!}{
\begin{tabular}{|c|c|c|c|c|c|c|c|c|c|c|c|c|c|c|c|c|}
\hline \multirow{2}{*}{ Algorithm } & Baseline & \multicolumn{1}{c|}{ NPOS } & \multicolumn{1}{c|}{ RoSAS } & \multicolumn{1}{c|}{ Mixup } & \multicolumn{1}{c|}{ Cutout } & \multicolumn{1}{c|}{ CutMix } & \multicolumn{1}{c|}{ GN } & \multicolumn{1}{c|}{ NNG-Mix } \\
\cline { 2 - 17 } & $1 \%$  & $10 \times$& $10 \times$  & $10 \times$  & $10 \times$  & $10 \times$  & $10 \times$  & $10 \times$ \\
\hline DeepSAD & 0.692&  \hlf{0.716} & 0.686&0.697 &0.698 &0.700&0.695 &\hls{0.708} \\ 
\hline XGBOD & 0.725&0.747 &0.711 & 0.720&\hls{0.749}  &\hls{0.749}  &0.744&\hlf{0.750} \\
\hline MLP & 0.657&\hlf{0.727} & 0.497& 0.634& 0.628&\hls{0.682} &0.658  &0.675 \\
\hline FTTransformer &0.564 &0.629 & 0.630& 0.644  &0.619  &0.626 &\hls{0.646}  &\hlf{0.652} \\
\hline CatB & \hls{0.769}&0.763 &0.749 & 0.768 &\hlf{0.776} &0.764 & 0.757 & 0.758\\
\hline \textit{Average} &  0.681& \hlf{0.716}& 0.655 &  0.693&   0.694 &   0.704&   0.700  &   \hls{0.709}   \\
\hline
\end{tabular}
}
\caption{Results on CV datasets with $1\%$ available labeled anomalies and $10 \times$ pseudo-anomaly generation. The best results are in \hlf{red} and the second best results are in \hls{blue}.}
\label{tab:CV-0.01-10-new}
\end{table*}
\begin{table*}[t!]
\centering
\resizebox{0.7\linewidth}{!}{
\begin{tabular}{|c|c|c|c|c|c|c|c|c|c|c|c|c|c|c|c|c|}
\hline \multirow{2}{*}{ Algorithm } & Baseline  & \multicolumn{1}{c|}{ NPOS } & \multicolumn{1}{c|}{ RoSAS } & \multicolumn{1}{c|}{ Mixup } & \multicolumn{1}{c|}{ Cutout } & \multicolumn{1}{c|}{ CutMix } & \multicolumn{1}{c|}{ GN } & \multicolumn{1}{c|}{ NNG-Mix } \\
\cline { 2 - 17 } & $1 \%$  & $10 \times$ & $10 \times$ & $10 \times$  & $10 \times$  & $10 \times$  & $10 \times$  & $10 \times$ \\
\hline DeepSAD & 0.561& 0.573& 0.584 &0.581 &\hls{0.582} & 0.581 &0.578 &\hlf{0.586} \\
\hline XGBOD &0.604 & 0.593 &0.589 &0.592 &\hls{0.627} &0.608 &0.617 &\hlf{0.631} \\ 
\hline MLP &0.586 &0.584& 0.644 &0.654 &\hls{0.664}&0.660 &0.659 & \hlf{0.666} \\ 
\hline FTTransformer & 0.495& 0.498&0.477&  0.494 &0.476 &\hls{0.497}  &0.487  &\hlf{0.507} \\ 
\hline CatB &0.603 &0.599&0.603&  0.606 &0.606 &0.604  &\hls{0.611}& \hlf{0.612}\\
\hline \textit{Average} & 0.570& 0.569& 0.579&  0.585&  \hls{0.591}&   0.590&    0.590&  \hlf{0.600}  \\
\hline
\end{tabular}
}
\caption{Results on NLP datasets with $1\%$ available labeled anomalies and $10 \times$ pseudo-anomaly generation. The best results are in \hlf{red} and the second best results are in \hls{blue}.}
\label{tab:NLP-0.01-10-new}
\end{table*}

\noindent\textbf{Generating $10 \times$ pseudo-anomalies using $1\%$ labeled anomalies.}
We commence our experiments in a highly challenging scenario, where only a mere $1\%$ of labeled anomalies is available. These labeled data, combined with unlabeled data, are used to generate $10 \times$ pseudo-anomalies employing various pseudo-anomaly generation algorithms. Subsequently, we train different semi-supervised and supervised algorithms on both original and generated data. In the Baseline scenario, the training solely relies on the original data containing $1\%$ labeled anomalies and unlabeled data, without integrating pseudo-anomalies. Results from \cref{tab:cls-0.01-10-new} to \cref{tab:NLP-0.01-10-new} present the outcomes across Classical, CV, and NLP datasets. For each dataset type, the reported metric is the mean AUCROC across all datasets within that type. Notably, our proposed NNG-Mix consistently demonstrates competitive performances across all dataset types, ranking as the best or the second best in most cases. 

An interesting observation is that almost all pseudo-anomaly generation algorithms demonstrate the ability to enhance anomaly detection performance by introducing an augmented set of pseudo-anomalies for training. When compared to the Baseline setup trained solely on the original data, leveraging the pseudo-anomalies generated by NNG-Mix yields significant benefits for all evaluated semi-supervised and supervised algorithms across various dataset types in most cases. Specifically, for semi-supervised algorithms, there are maximum improvements of $2.5\%$, $2.5\%$, $2.7\%$ observed in Classical, CV, and NLP datasets, respectively. For supervised algorithms, there are maximum improvements of $16.4\%$, $8.8\%$, $8.0\%$ observed in Classical, CV, and NLP datasets, respectively. On Classical and NLP datasets, MLP benefits the most from NNG-Mix, while FTTransformer exhibits the most improvement on CV datasets. Additionally, CutMix~\cite{yun2019cutmix}, a conventional data augmentation method in computer vision, showcases competitive performances on CV datasets, with an average improvement of $2.2\%$. Similarly, Cutout~\cite{devries2017cutout}, resembling random masking~\cite{devlin2018bert} in natural language processing, also demonstrates competitive results on NLP datasets, with an average improvement of $2.1\%$. NPOS~\cite{tao2023nonparametric} demonstrates competitive performance on CV datasets, yet shows limited improvements on Classical and NLP datasets. NNG-Mix is the \textit{only} method that consistently delivers  substantial performance improvements across  all  dataset types.

\noindent\textbf{Influences of the number of generated pseudo-anomalies.} We then analyze the performance of the proposed algorithm with different numbers of generated pseudo-anomalies on Classical datasets. Specifically, we generate $1 \times$, $5 \times$, and $10 \times$ the number of anomalies as pseudo-anomalies using various anomaly generation algorithms. Subsequently, we train different semi-supervised and supervised algorithms on datasets that combine the original and the generated data samples. As depicted in \cref{tab:cls-0.01}, for most pseudo-anomaly generation algorithms, increasing the number of pseudo-anomalies tends to yield improved results.
Generating $1 \times$ pseudo-anomalies typically results in some improvements compared to the Baseline in most cases. The results become notably better when generating $5 \times$ pseudo-anomalies, with the best performance achieved in the $10 \times$ scenario.

\begin{table*}[ht!]
\centering
\resizebox{\linewidth}{!}{
\begin{tabular}{|c|c|c|c|c|c|c|c|c|c|c|c|c|c|c|c|c|}
\hline \multirow{2}{*}{ Algorithm } & Baseline & \multicolumn{3}{c|}{ Mixup } & \multicolumn{3}{c|}{ Cutout } & \multicolumn{3}{c|}{ CutMix } & \multicolumn{3}{c|}{ GN } & \multicolumn{3}{c|}{ NNG-Mix } \\
\cline { 2 - 17 } & $1 \%$ & $1 \times$ & $5 \times$ & $10 \times$ & $1 \times$ & $5 \times$ & $10 \times$ & $1 \times$ & $5 \times$ & $10 \times$ & $1 \times$ & $5 \times$ & $10 \times$ & $1 \times$ & $5 \times$ & $10 \times$ \\
\hline DeepSAD & 0.766& 0.768&0.779 &\hls{0.786} & 0.765&0.774 &0.780 &0.765 &0.775 &0.782 &0.766 &0.772 &0.782 &0.768 &0.779 &\hlf{0.791} \\
\hline XGBOD &0.815 &0.795 &0.790 &0.785 &0.799 &0.796 &0.798 & 0.796&0.805 &0.816 & 0.817&0.832 &\hls{0.837} &0.820 &\hlf{0.838} &\hlf{0.838} \\
\hline MLP & 0.600& 0.646& 0.745& 0.758&0.647 & 0.733&0.748 &0.649 &0.739 &\hls{0.763} &0.666 &0.743 &0.751 &0.634 &0.743 &\hlf{0.764} \\
\hline FTTransformer &0.719 & 0.730&0.769 &\hls{0.781} &0.722 &0.754 &0.757 &0.720 &0.757 &0.778 & 0.721& 0.754& 0.763&0.721&0.762 &\hlf{0.783} \\
\hline CatB & 0.851&0.848 & 0.848&0.846 & 0.842& 0.839& 0.836& 0.847& 0.840&0.844 & 0.854&0.853 &\hls{0.859} & 0.851& 0.854&\hlf{0.861} \\
\hline \textit{Average} & 0.750 & 0.757 & 0.786 & 0.791 & 0.755 & 0.779 & 0.784 & 0.755 & 0.783 & 0.797 & 0.765 & 0.791 & \hls{0.798} & 0.759 & 0.795 & \hlf{0.807}\\
\hline
\end{tabular}
}
\caption{Results on Classical datasets with $1\%$ available labeled anomalies and $1 \times$, $5 \times$, $10 \times$ pseudo-anomaly generation. The best results are in \hlf{red} and the second best results are in \hls{blue}.}
\label{tab:cls-0.01}
\end{table*}

\begin{table*}[t!]
\centering
\resizebox{0.85\linewidth}{!}{
\begin{tabular}{|c|c|c|c|c|c|c|c|c|c|c|c|c|c|c|c|c|}
\hline \multirow{2}{*}{ Algorithm } & Baseline & \multicolumn{3}{c|}{ NNG-mix } & \multicolumn{1}{c|}{ Baseline } & \multicolumn{3}{c|}{ NNG-mix } & \multicolumn{1}{c|}{Baseline } & \multicolumn{3}{c|}{ NNG-mix } \\
\cline { 2 - 17 } & $\mathbf{1 \%}$ & $1 \times$ & $5 \times$ & $10 \times$ & $\mathbf{5 \%}$ & $1 \times$ & $5 \times$ & $10 \times$ & $\mathbf{10 \%}$ & $1 \times$ & $5 \times$ & $10 \times$ \\
\hline DeepSAD & 0.766  &0.768& 0.779 &\hlf{0.791} &0.817 & 0.828 &0.848 &\hlf{0.850} & 0.853&0.865 &0.880 & \hlf{0.888}  \\
\hline XGBOD &0.815 &0.820 &\hlf{0.838} &\hlf{0.838} & 0.868& 0.877 & \hlf{0.886} & \hlf{0.886} &0.897 & 0.912& 0.911 &  \hlf{0.914}  \\
\hline MLP & 0.600&0.634 &0.743 &\hlf{0.764} & 0.735&0.775 &0.829 &\hlf{0.853}& 0.791& 0.838 &0.879 &\hlf{0.891} \\
\hline FTTransformer &0.719 &0.721&0.762 &\hlf{0.783} & 0.749& 0.786 & 0.816 & \hlf{0.853} &0.787 & 0.810  & 0.859 &  \hlf{0.870} \\
\hline CatB & 0.851& 0.851& 0.854&\hlf{0.861}&0.891& 0.894&  0.899&\hlf{0.901}  &0.916 &0.921 &\hlf{0.923} & 0.922 \\
\hline \textit{Average} & 0.750 & 0.759 & 0.795 & \hlf{0.807} & 0.812 & 0.832 & 0.856 &\hlf{0.869} & 0.849 & 0.869 & 0.890 & \hlf{0.897} \\
\hline
\end{tabular}
}
\caption{Results on Classical datasets with $1\%$, $5\%$, $10\%$ available labeled anomalies, and $1 \times$, $5 \times$, $10 \times$ pseudo-anomaly generation based on the number of available labeled anomalies using NNG-Mix. The best results for each baseline are in \hlf{red}.}
\label{tab:cls-NNG-Mix}
\end{table*}
\noindent\textbf{Influences of the number of labeled anomalies.} Here, we analyze the performance under the availability of different amounts of labeled anomalies on Classical datasets using NNG-Mix. With an increase in the number of labeled anomalies, both the semi-supervised and supervised anomaly detection Baselines exhibit improved performances. The results, presented in \cref{tab:cls-NNG-Mix}, indicate that NNG-Mix can further enhance the Baselines with different ratios of labeled anomalies by generating more pseudo-anomalies for training, confirming the effectiveness of our proposed approach.
For instance, applying NNG-Mix to the DeepSAD model and generating $10$ times the number of pseudo-anomalies results in an increase of AUCROC scores by $2.5\%$, $3.3\%$, and $3.5\%$, respectively, in scenarios with $1\%$, $5\%$, and $10\%$ of labeled anomalies. This observation highlights the consistent performance enhancement of the DeepSAD model when integrated with generated pseudo-anomalies, even in situations where a sufficient number of labeled anomalies is already available. This indicates that NNG-Mix improves the generalization of SOTA semi-supervised and supervised AD algorithms.

\begin{figure}[ht]
  \centering
  \begin{subfigure}{0.5\textwidth}
  \centering
    \includegraphics[width=0.8\textwidth]{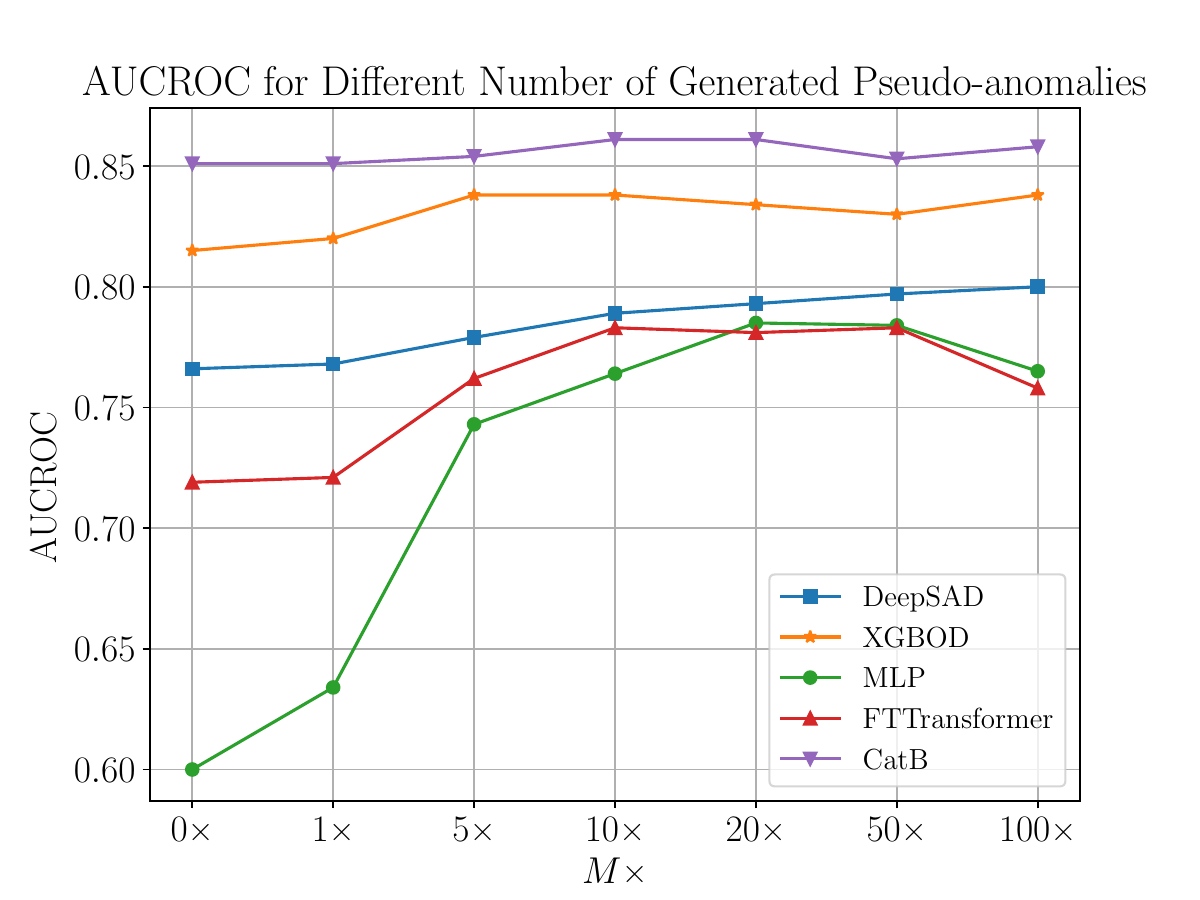}
    \caption{Number of Generated Pseudo-anomalies}
  \end{subfigure}
  \begin{subfigure}{0.5\textwidth}
  \centering
    \includegraphics[width=0.8\textwidth]{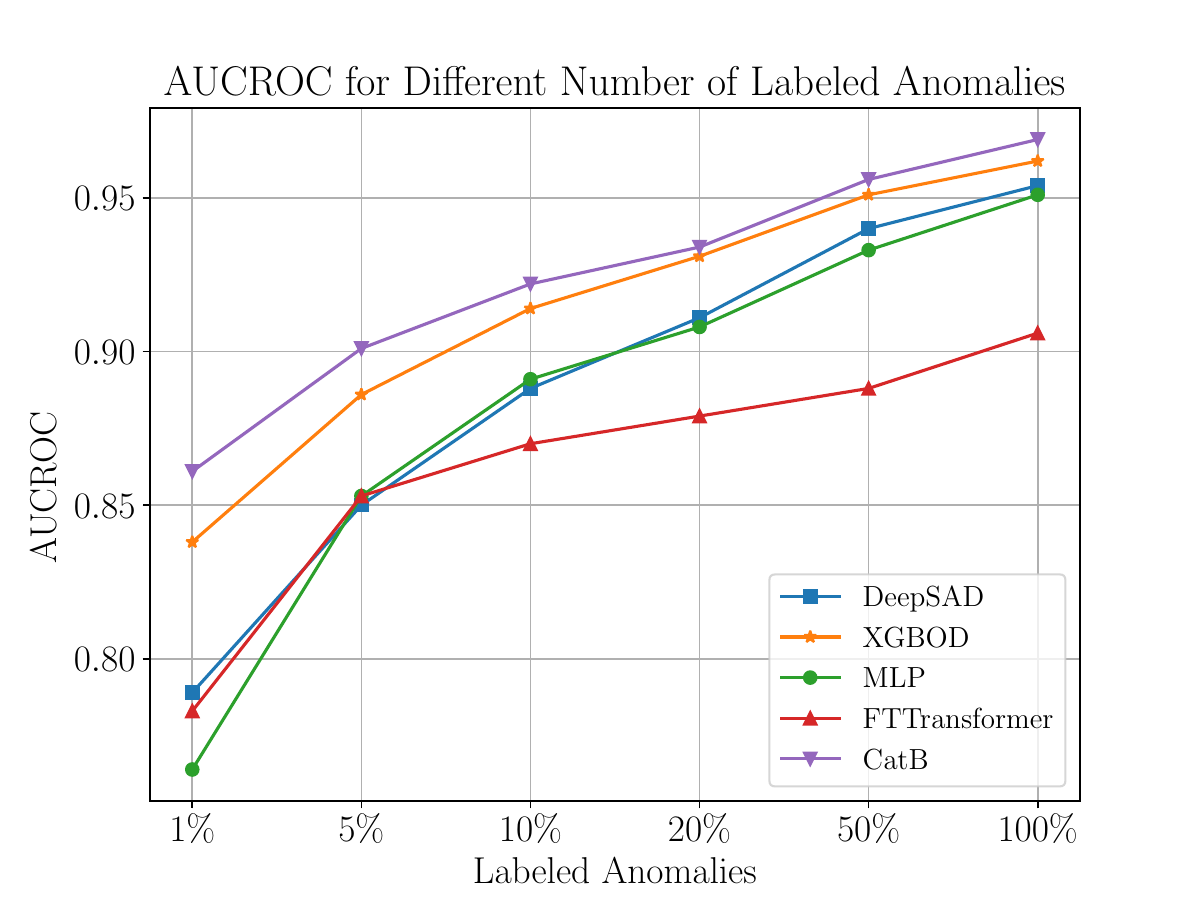}
    \caption{Number of Labeled Anomalies}
  \end{subfigure}
  \caption{Ablations on different numbers of generated pseudo-anomalies and labeled anomalies. Generating more pseudo-anomalies brings performance improvement in general but tends to plateau when $M$ exceeds $10$. With the availability of more labeled anomalies, the performances of all algorithms are improved significantly. }
   \label{fig:ng}
\end{figure}

\noindent\textbf{Detailed results on $47$ Classical datasets.} \cref{tab:cls-0.1-detail} presents detailed results from applying  DeepSAD with $10\%$ available labeled anomalies on $47$ Classical datasets, supplemented by pseudo-anomaly generation at levels  $1 \times$, $5 \times$, $10 \times$  using NNG-Mix. Our NNG-Mix method enhances  baseline performance across  the majority of these  datasets, even with just  $1 \times$ pseudo-anomaly generation. Notably, with $10 \times$ pseudo-anomaly generation, our method demonstrates  improvements in $41$ out of $47$ datasets, achieving a maximum improvement of $22.9\%$ and an average improvement of $3.5\%$. These findings highlight  the significant impact  of generating additional pseudo-anomalies in enhancing anomaly detection performance. 

\begin{table*}[ht!]
\centering
\resizebox{0.85\linewidth}{!}{
\begin{tabular}{|c|c|c|c|c|c|c|c|c|c|c|c|c|c|}
\hline \multirow{2}{*}{ Algorithm } & Baseline & \multicolumn{3}{c|}{ w/o NN and GN} & \multicolumn{3}{c|}{ w/o NN } & \multicolumn{3}{c|}{ w/o GN } & \multicolumn{3}{c|}{ NNG-Mix } \\
\cline { 2 - 14 } & $1 \%$ &  $1 \times$ & $5 \times$ & $10 \times$ &  $1 \times$ & $5 \times$ & $10 \times$ & $1 \times$ & $5 \times$ & $10 \times$ & $1 \times$ & $5 \times$ & $10 \times$ \\
\hline DeepSAD & 0.766&0.766 &0.770 &0.769 &0.767 &0.765 &0.779 &0.767 &0.779 & 0.790   &0.768 &0.779 &\hlf{0.791} \\
\hline XGBOD &0.815&0.805 &0.819 & 0.796& 0.820&0.824 &0.826 &0.797 &0.788 &0.795   &0.820 &0.838 &\hlf{0.838}  \\
\hline MLP & 0.600& 0.629&0.687 &0.724 &0.625 & 0.702& 0.706&0.637 & 0.738& 0.763 &0.634 &0.743 &\hlf{0.764} \\
\hline FTTransformer &0.719 &0.722 &0.760 &0.746 & 0.730&0.765 &0.750 &0.743 &0.760 &0.765  &0.721& 0.762 &\hlf{0.783}  \\
\hline CatB & 0.851&0.842 & 0.825&0.802 &0.845 &0.838 &0.843 &0.846 &0.842 &0.840  & 0.851& 0.854 &\hlf{0.861} \\
\hline \textit{Average} & 0.750 & 0.753 & 0.772 & 0.767 & 0.757 & 0.779 & 0.781 & 0.758 & 0.781 & 0.791 & 0.759 & 0.795 & \hlf{0.807}\\
\hline
\end{tabular}
}
\caption{Ablations on Classical datasets with $1\%$ available labeled anomalies and $1 \times$, $5 \times$, $10 \times$ pseudo-anomaly generation using NNG-Mix. The best results for each baseline are in \hlf{red}.}
\label{tab:cls-0.01-abla}
\end{table*}
\begin{table}[ht!]
    \centering
    \begin{tabular}{lcccc}
        \toprule
        Dataset & Baseline & $1 \times$ & $5 \times$ & $10 \times$ \\
        \midrule
        landsat & 0.896 & 0.892 & 0.904 & 0.898 (\textcolor{blue}{+0.2\%}) \\
        wine & 0.813 & 0.894 & 0.990 & 0.999 (\textcolor{blue}{+18.6\%}) \\
        shuttle & 0.996 & 0.996 & 0.997 & 0.996 (\textcolor{blue}{+0.0\%}) \\
        skin & 0.998 & 0.998 & 0.999 & 0.999 (\textcolor{blue}{+0.1\%}) \\
        Cardiotocography & 0.860 & 0.870 & 0.889 & 0.883 (\textcolor{blue}{+2.3\%}) \\
        Waveform & 0.773 & 0.784 & 0.853 & 0.864 (\textcolor{blue}{+9.1\%}) \\
        glass & 0.941 & 0.961 & 0.979 & 0.989 (\textcolor{blue}{+4.8\%}) \\
        breastw & 0.957 & 0.967 & 0.980 & 0.979 (\textcolor{blue}{+2.2\%}) \\
        yeast & 0.652 & 0.673 & 0.675 & 0.683 (\textcolor{blue}{+3.1\%}) \\
        PageBlocks & 0.955 & 0.952 & 0.943 & 0.947 (\textcolor{red}{-0.8\%}) \\
        satellite & 0.917 & 0.916 & 0.921 & 0.923 (\textcolor{blue}{+0.6\%}) \\
        InternetAds & 0.740 & 0.748 & 0.771 & 0.783 (\textcolor{blue}{+4.3\%}) \\
        campaign & 0.650 & 0.658 & 0.686 & 0.721 (\textcolor{blue}{+7.1\%}) \\
        speech & 0.480 & 0.478 & 0.486 & 0.503 (\textcolor{blue}{+2.3\%}) \\
        annthyroid & 0.898 & 0.895 & 0.900 & 0.908 (\textcolor{blue}{+1.0\%}) \\
        Lymphography & 0.917 & 0.943 & 0.952 & 0.983 (\textcolor{blue}{+6.6\%}) \\
        census & 0.795 & 0.797 & 0.796 & 0.788 (\textcolor{red}{-0.7\%}) \\
        Hepatitis & 0.848 & 0.861 & 0.865 & 0.856 (\textcolor{blue}{+0.8\%}) \\
        thyroid & 0.980 & 0.988 & 0.985 & 0.984 (\textcolor{blue}{+0.4\%}) \\
        Stamps & 0.740 & 0.782 & 0.878 & 0.924 (\textcolor{blue}{+18.4\%}) \\
        mammography & 0.939 & 0.942 & 0.937 & 0.940 (\textcolor{blue}{+0.1\%}) \\
        magic.gamma & 0.819 & 0.822 & 0.821 & 0.811 (\textcolor{red}{-0.8\%}) \\
        vowels & 0.600 & 0.687 & 0.748 & 0.829 (\textcolor{blue}{+22.9\%}) \\
        WDBC & 0.819 & 0.815 & 0.748 & 0.820 (\textcolor{blue}{+0.1\%}) \\
        SpamBase & 0.789 & 0.801 & 0.839 & 0.844 (\textcolor{blue}{+5.5\%}) \\
        mnist & 0.945 & 0.948 & 0.960 & 0.958 (\textcolor{blue}{+1.3\%}) \\
        cover & 0.929 & 0.996 & 0.998 & 0.998 (\textcolor{blue}{+6.9\%}) \\
        ALOI & 0.590 & 0.603 & 0.606 & 0.612 (\textcolor{blue}{+2.2\%}) \\
        celeba & 0.816 & 0.832 & 0.879 & 0.868 (\textcolor{blue}{+5.2\%}) \\
        Ionosphere & 0.884 & 0.898 & 0.904 & 0.898 (\textcolor{blue}{+1.4\%}) \\
        backdoor & 0.964 & 0.973 & 0.977 & 0.977 (\textcolor{blue}{+1.3\%}) \\
        Wilt & 0.828 & 0.836 & 0.883 & 0.897 (\textcolor{blue}{+6.9\%}) \\
        pendigits & 0.994 & 0.998 & 0.998 & 0.999 (\textcolor{blue}{+0.5\%}) \\
        vertebral & 0.747 & 0.792 & 0.877 & 0.935 (\textcolor{blue}{+18.8\%}) \\
        WBC & 0.943 & 0.933 & 0.951 & 0.949 (\textcolor{blue}{+0.6\%}) \\
        musk & 0.999 & 1.000 & 0.999 & 1.000 (\textcolor{blue}{+0.1\%}) \\
        WPBC & 0.637 & 0.645 & 0.636 & 0.630 (\textcolor{red}{-0.7\%}) \\
        donors & 1.000 & 1.000 & 1.000 & 1.000 (\textcolor{blue}{+0.0\%}) \\
        smtp & 1.000 & 1.000 & 1.000 & 1.000 (\textcolor{blue}{+0.0\%}) \\
        cardio & 0.914 & 0.951 & 0.964 & 0.969 (\textcolor{blue}{+5.5\%}) \\
        http & 0.998 & 0.999 & 0.999 & 1.000 (\textcolor{blue}{+0.2\%}) \\
        optdigits & 0.992 & 0.995 & 0.992 & 0.997 (\textcolor{blue}{+0.5\%}) \\
        letter & 0.761 & 0.764 & 0.815 & 0.812 (\textcolor{blue}{+5.1\%}) \\
        satimage-2 & 0.986 & 0.987 & 0.991 & 0.989 (\textcolor{blue}{+0.3\%}) \\
        fraud & 0.976 & 0.983 & 0.963 & 0.957 (\textcolor{red}{-1.9\%}) \\
        fault & 0.731 & 0.731 & 0.713 & 0.708 (\textcolor{red}{-2.3\%}) \\
        Pima & 0.665 & 0.695 & 0.717 & 0.722 (\textcolor{blue}{+5.7\%}) \\
        \textit{Average} & 0.853 & 0.865 & 0.880 & 0.888 (\textcolor{blue}{+3.5\%}) \\
        \bottomrule
    \end{tabular}
\caption{Detailed results on $47$ Classical datasets using DeepSAD with $10\%$ available labeled anomalies and $1 \times$, $5 \times$, $10 \times$ pseudo-anomaly generation using NNG-Mix. When generating $10 \times$ pseudo-anomalies, our method shows improvements in $41$ out of $47$ datasets, with a maximum improvement of $22.9\%$ and an average improvement of $3.5\%$.}   
\label{tab:cls-0.1-detail}
\end{table}

\subsection{Ablations}
\subsubsection{Ablations on Each Module in NNG-Mix}
We conducted extensive ablation studies to investigate the role of each module within NNG-Mix on the Classical datasets, considering scenarios with $1\%$ available labeled anomalies and $1 \times$, $5 \times$, $10 \times$ pseudo-anomaly generation, as shown in~\cref{tab:cls-0.01-abla}.
NNG-Mix without Nearest Neighbor (NN) and Gaussian noises (GN) corresponds to a special case of Mixup, where samples are drawn from both the labeled anomaly data $\mathcal{A}$ and unlabeled data $\mathcal{H}$ for mixing, with all considered as anomalies. This random mixing of anomalies and unlabeled data can inadvertently generate anomaly data points falling within the distribution of normal data, resulting in diminished performance (e.g., $1.9\%$ for XGBOD and $4.9\%$ for CatB) compared to the Baseline setup. Even when Gaussian noise is introduced before applying Mixup, omitting Nearest Neighbor results in underwhelming performance (e.g., only $1.3\%$ increasement for DeepSAD and even $0.8\%$ decreasement for CatB). It becomes evident that the primary enhancement in performance is achieved by incorporating the Nearest Neighbor search without the use of Gaussian noise. Ultimately, the best performance is observed when Gaussian noise is added to enlarge the potential space for pseudo-anomaly generation.

\subsubsection{Ablations on Different Numbers of Generated Pseudo-anomalies}
In this section, we investigate the impact of the quantity of generated pseudo-anomalies on the final AD performance. We employ NNG-Mix to create $M \times$ number of pseudo-anomalies, where $M$ varies from $1, 5, 10, 20, 50, 100$. Generally, the performance improves with an increasing number of generated pseudo-anomalies, as illustrated in~\cref{fig:ng} (a). However, the performance tends to plateau when $M$ exceeds $10$. One potential reason is the increased probability of injecting noise samples when $M$ increases. Although NNG-Mix can generate pseudo-anomalies effectively and avoid undesired samples as much as possible, it may still produce some noise samples when $M$ is too large and adversely impact the performance of anomaly detection.

\subsubsection{Ablations on Different Quantities of Labeled Anomalies}
Here we investigate the influence of the quantity of labeled anomalies on the final AD performance. We use NNG-Mix to generate $10 \times$ the number of pseudo-anomalies, considering varying proportions of available labeled anomalies: $1\%$, $5\%$, $10\%$, $20\%$, $50\%$, $100\%$. The performance of all algorithms improves significantly with an increased number of labeled anomalies, emphasizing the importance of supervision, as shown in~\cref{fig:ng} (b).

\begin{figure*}[t]
  \centering
  \begin{subfigure}{0.325\textwidth}
    \includegraphics[width=\textwidth]{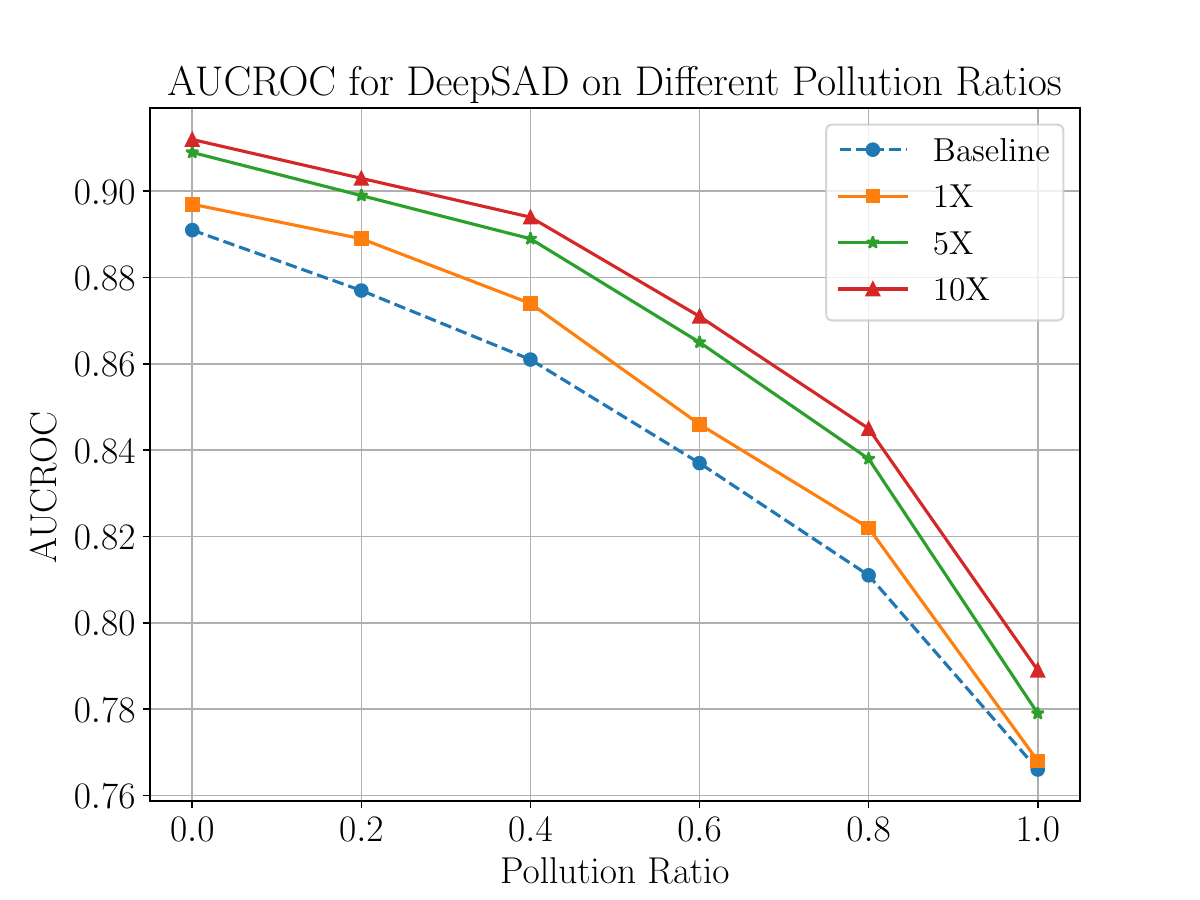}
    \caption{DeepSAD}
  \end{subfigure}
  \begin{subfigure}{0.325\textwidth}
    \includegraphics[width=\textwidth]{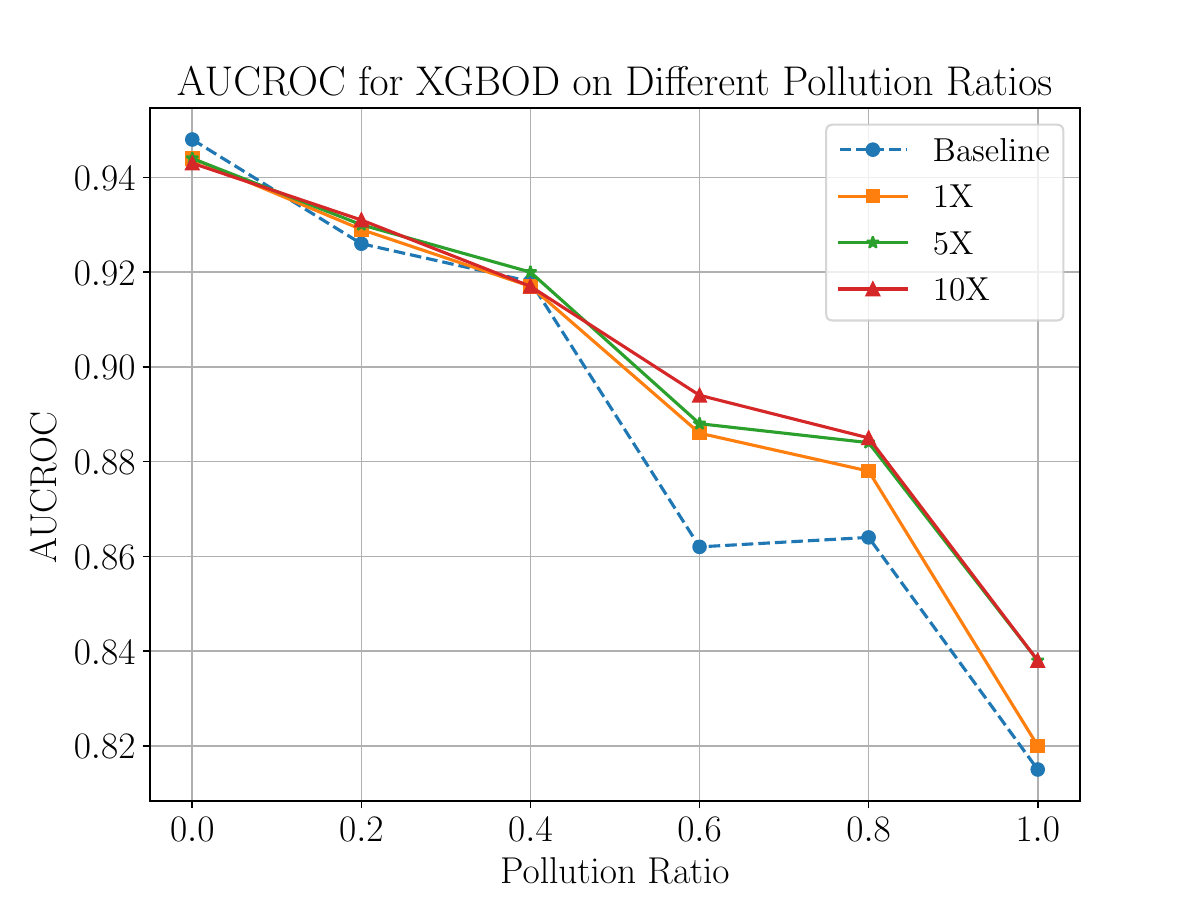}
    \caption{XGBOD}
  \end{subfigure}
  \begin{subfigure}{0.325\textwidth}
    \includegraphics[width=\textwidth]{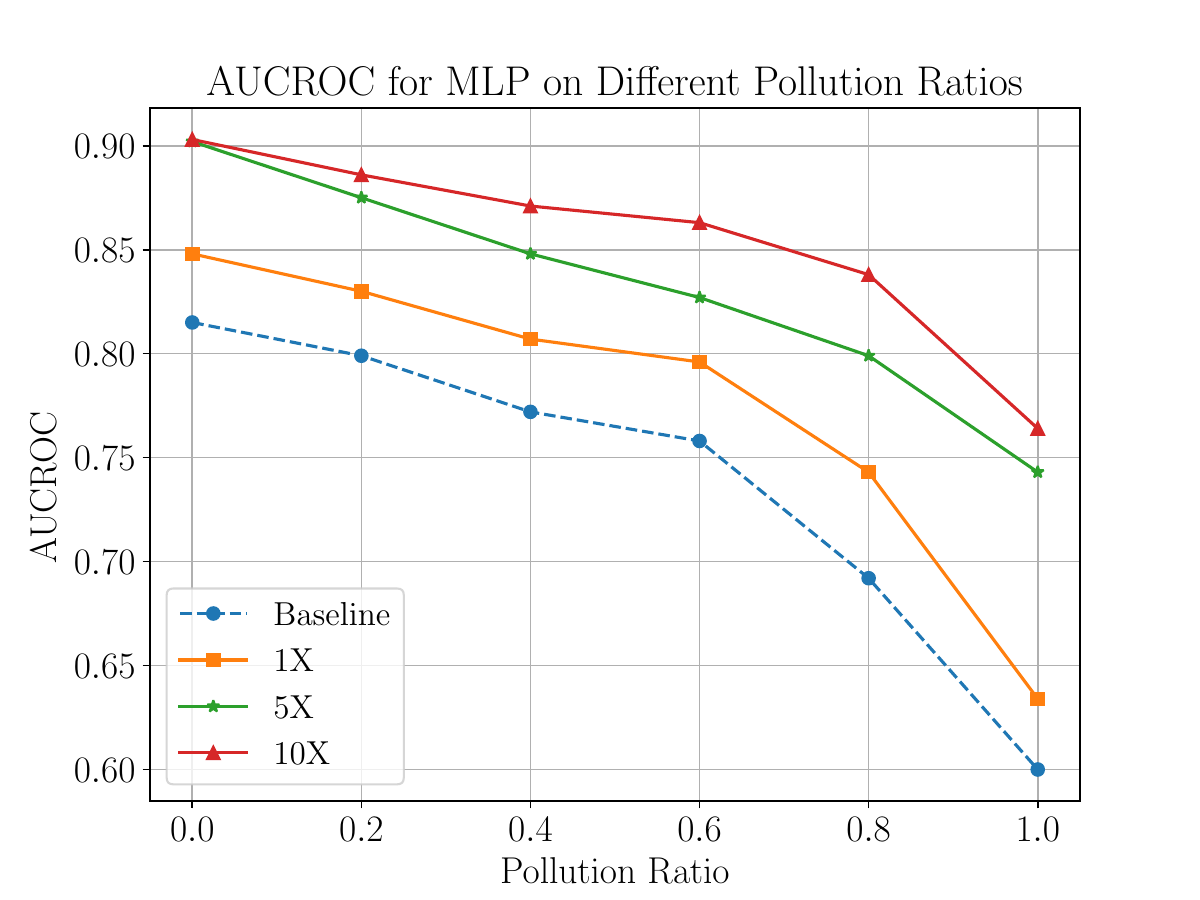}
    \caption{MLP}
  \end{subfigure}
  \begin{subfigure}{0.325\textwidth}
    \includegraphics[width=\textwidth]{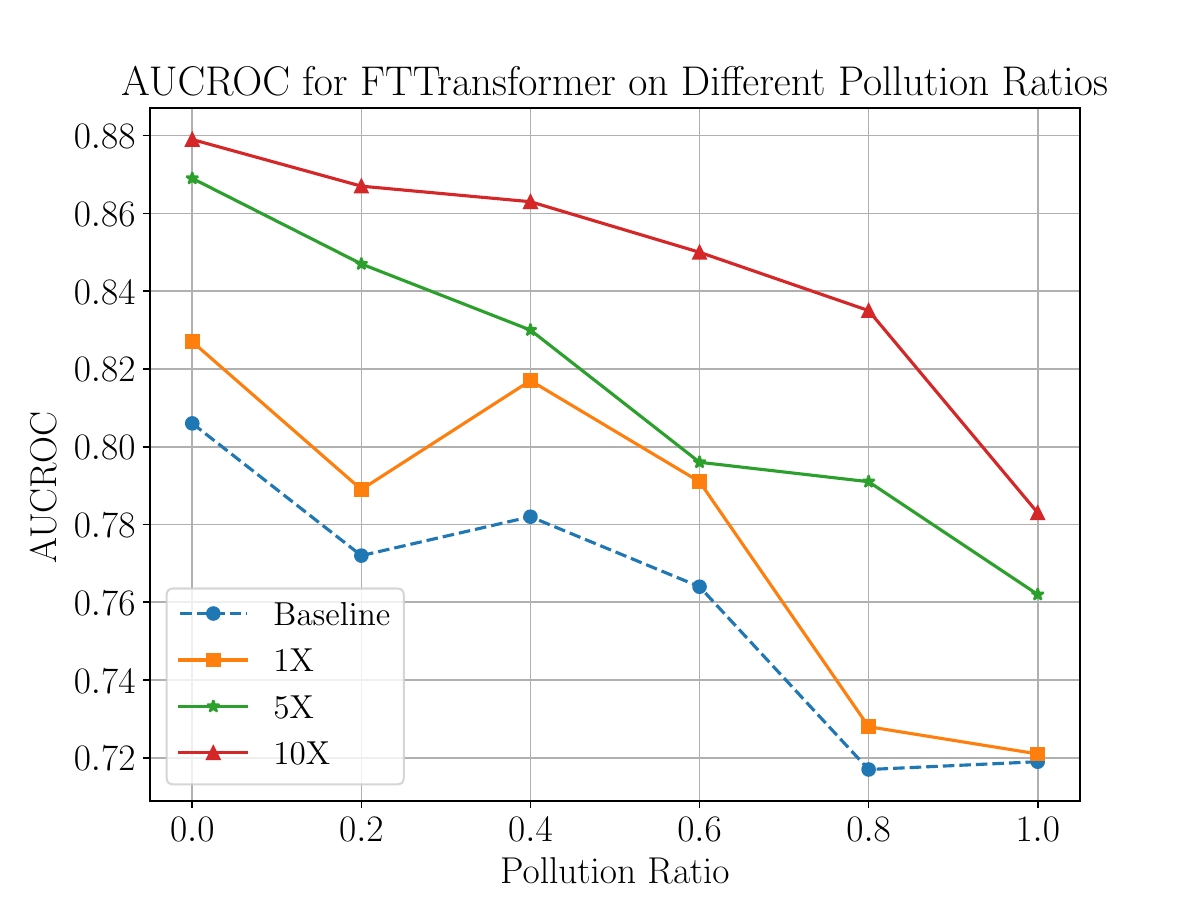}
    \caption{FTTransformer}
  \end{subfigure}
  \begin{subfigure}{0.325\textwidth}
    \includegraphics[width=\textwidth]{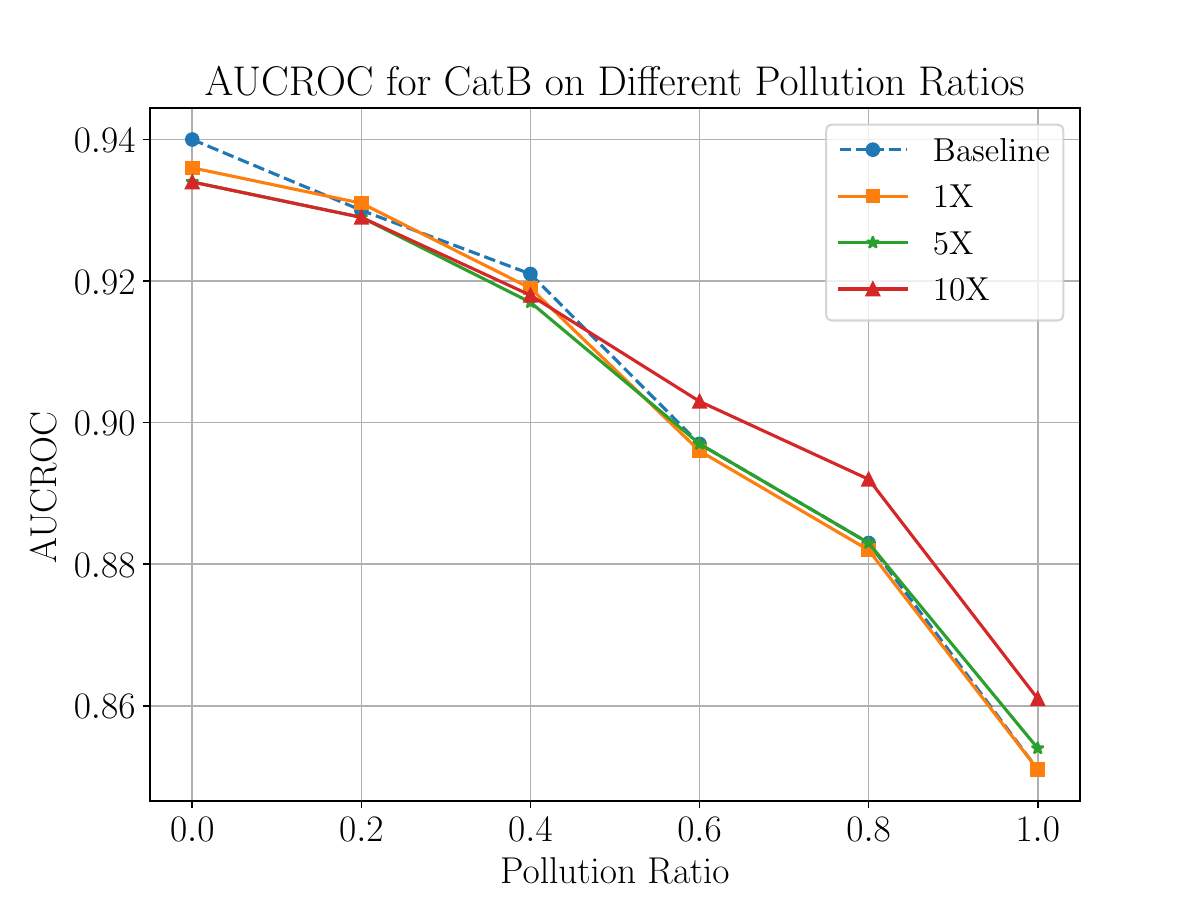}
    \caption{CatB}
  \end{subfigure}
  \caption{AUCROC with different pollution ratios. Lower pollution ratios consistently lead to performance improvements across all algorithms. Notably, for DeepSAD, MLP, and FTTransformer, NNG-Mix substantially enhances the Baseline setups, regardless of the pollution ratio. In contrast, for XGBOD and CatB, when the pollution ratio is low, the Baselines already exhibit superior performance, and the introduction of additional pseudo-anomalies does not yield significant benefits. NNG-Mix demonstrates its efficacy primarily when the pollution ratio surpasses the $0.6$ threshold.}
   \label{fig:pr}
\end{figure*}

\subsubsection{Ablations on Different Pollution Ratios}
In this section, we explore the robustness of NNG-Mix concerning various pollution ratios, denoted as $\boldsymbol{\gamma}$, within the unlabeled data $\mathcal{H}$, which means that we add different numbers of anomalies in the unlabeled data and treat them as normal data during pseudo-anomaly generation and training. To investigate this, we generate unlabeled data $\mathcal{H}$ by contaminating the normal data with different numbers of anomalies drawn from all available anomaly data, excluding the designated $1\%$ labeled anomalies. We set pollution ratios $\boldsymbol{\gamma}$ within a range of values, including $0.0$, $0.2$, $0.4$, $0.6$, $0.8$, and $1.0$, where $1.0$ represents the default setup used in all our previous experiments. This is also the default setup in ADBench and further AD papers. The results, illustrated in~\cref{fig:pr}, reveal that lower pollution ratios consistently lead to performance improvements across all algorithms. Notably, for DeepSAD, MLP, and FTTransformer, NNG-Mix substantially enhances the Baseline setups, regardless of the pollution ratio. In contrast, for XGBOD and CatB, when the pollution ratio is low, the Baselines already exhibit superior performance, and the introduction of additional pseudo-anomalies does not yield significant benefits. NNG-Mix demonstrates its efficacy primarily when the pollution ratio surpasses the $0.6$ threshold. An unexpected performance improvement for FTTransformer is observed with increased pollution ratios in certain cases. This phenomenon may be attributed to a unique characteristic of the FTTransformer algorithm. As demonstrated in~\cite{wei2021open}, learning with noisy labels can help the model converge to a flat minimum with superior stability. Nonetheless, the overall trend across all algorithms is a decline in performance with higher pollution ratios.

\subsubsection{Combination of Different Pseudo-Anomaly Generation Algorithms} Our NNG-Mix can be viewed  as a Nearest Neighbor variant of Mixup, augmented with Gaussian noise. To demonstrate the superiority of our method over other methods, we assessed  the performance of combining Gaussian noise with Cutout and CutMix. As shown in~\cref{tab:base_comb}, merely  combining different baseline algorithms typically  results in negligible improvements or even performance degradation. This underscores the effectiveness of NNG-Mix in generating pseudo-anomalies. 

\begin{table}[t!]
    \centering
    \begin{tabular}{ccc}
        \toprule
        Cutout+GN & CutMix+GN & NNG-Mix\\
        \midrule
          0.778 & 0.783 & 0.791\\
        \bottomrule
    \end{tabular}
\caption{Ablation on combinations of different pseudo-anomaly generation
algorithms. The experiments are on Classical datasets with $1\%$ available labeled anomalies and $10 \times$ pseudo-anomaly generation using DeepSAD and the AUCROC is reported.  } 
\label{tab:base_comb}
\end{table}

\subsubsection{Parameter Sensitivity}
\begin{figure*}[t]
  \centering
  \begin{subfigure}{0.325\textwidth}
    \includegraphics[width=\textwidth]{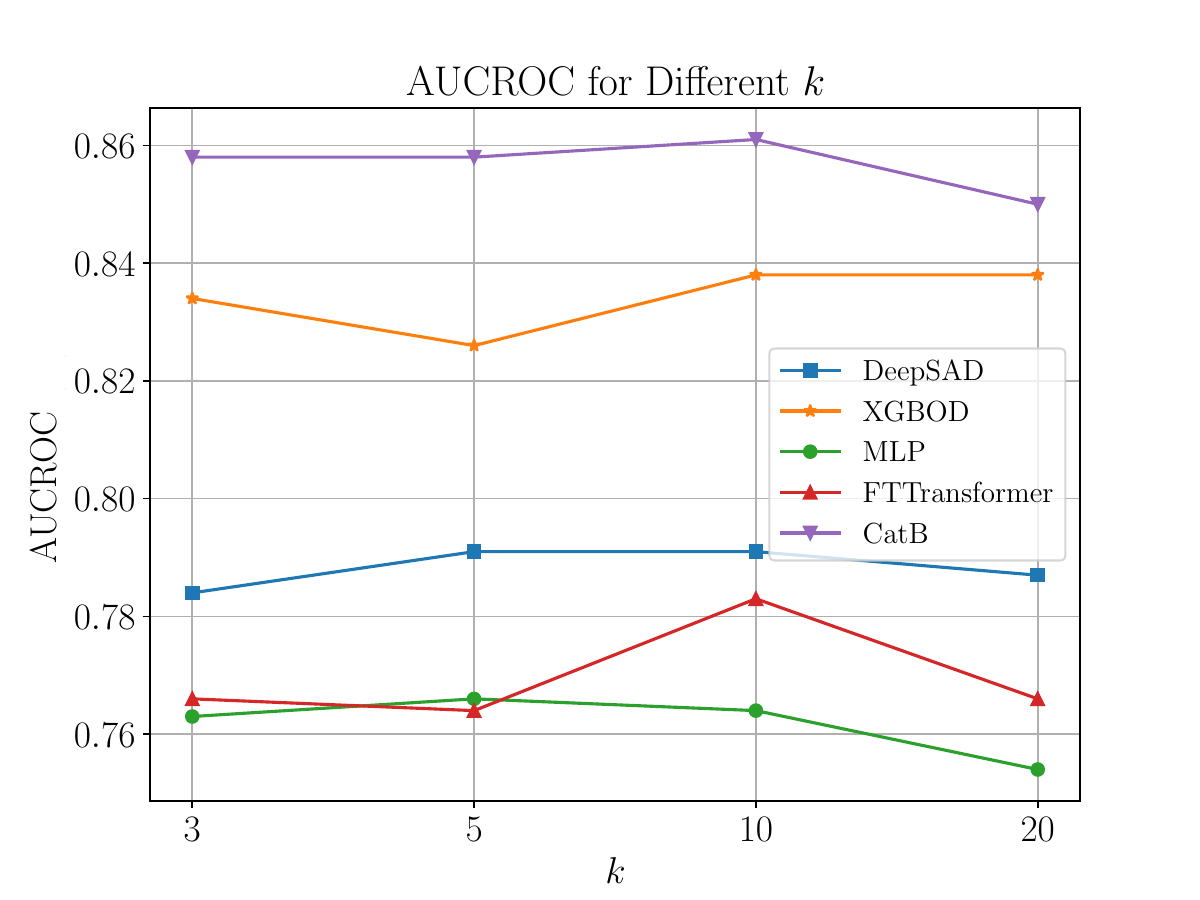}
    \caption{$k$}
  \end{subfigure}
  \begin{subfigure}{0.325\textwidth}
    \includegraphics[width=\textwidth]{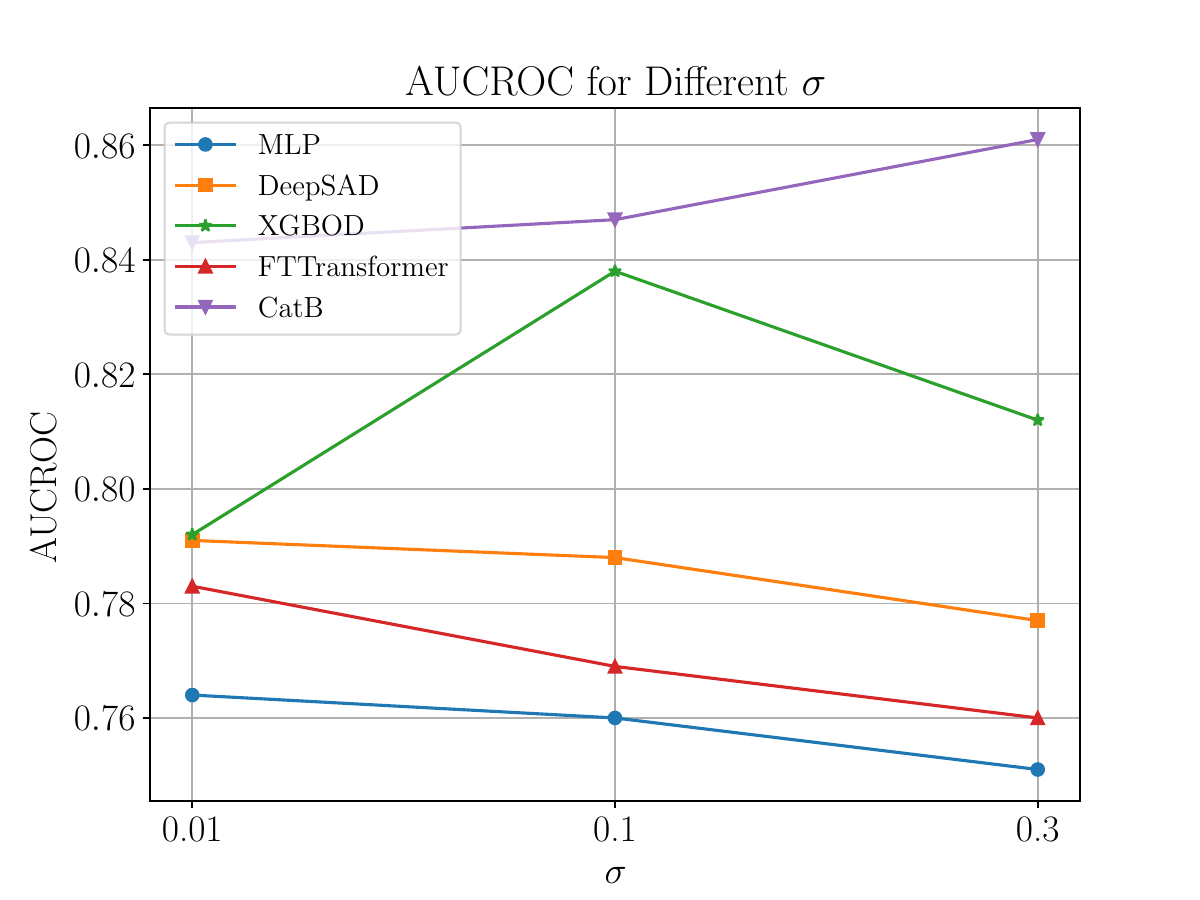}
    \caption{$\boldsymbol{\sigma}$}
  \end{subfigure}
  \begin{subfigure}{0.325\textwidth}
    \includegraphics[width=\textwidth]{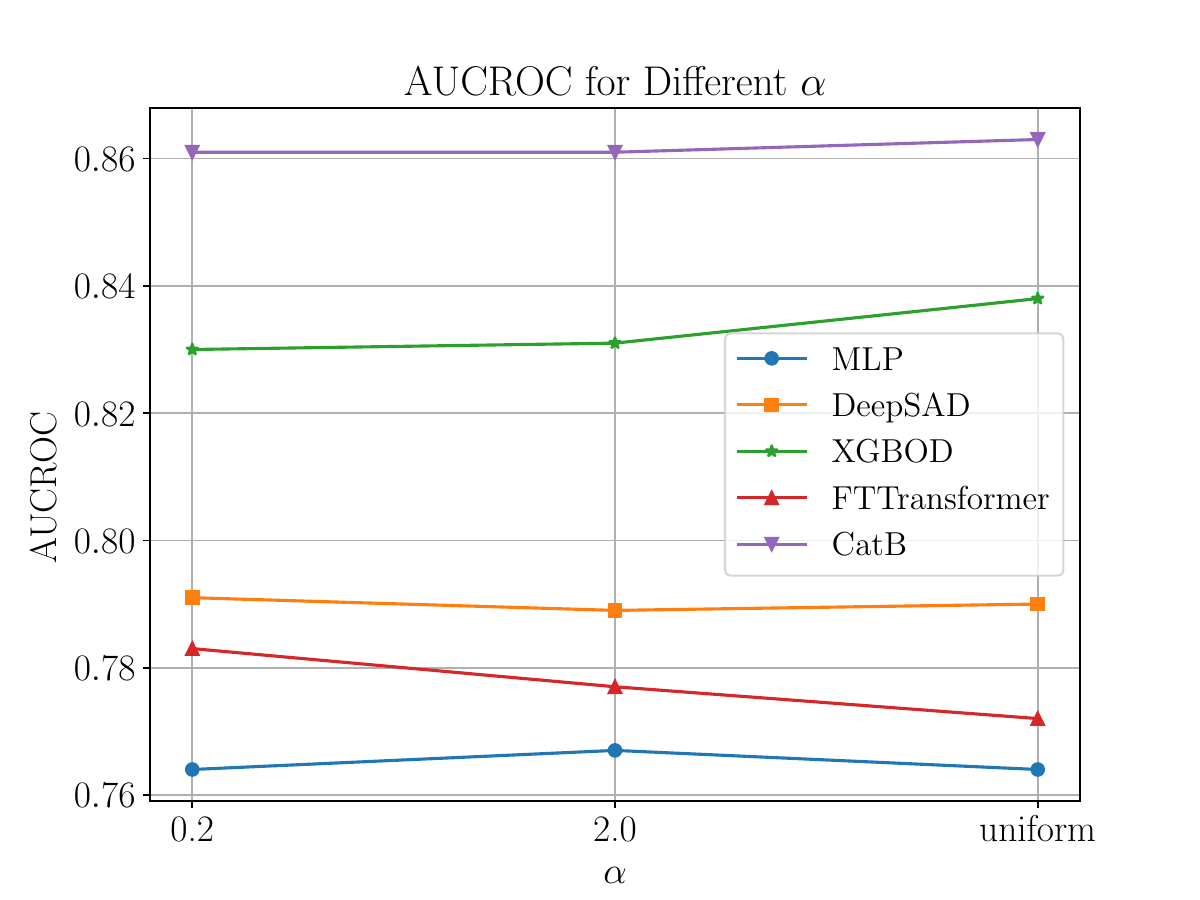}
    \caption{$\alpha$}
  \end{subfigure}
  \caption{Parameter Sensitivity. NNG-Mix demonstrates robustness across different parameter settings, displaying fluctuations of less than $2\%$ in most cases.}
   \label{fig:ps}
\end{figure*}

In this section, we investigate the parameter sensitivity of NNG-Mix on the Classical datasets, with $1\%$ available labeled anomalies and $10 \times$ pseudo-anomaly generation, as depicted in~\cref{fig:ps}. For the Nearest Neighbor parameter $k$, we test values of $3$, $5$, $10$, and $20$. Regarding the Gaussian noise parameter $\boldsymbol{\sigma}$, we consider choices of $0.01$, $0.1$, and $0.3$. Additionally, for the Mixup parameter $\alpha$, we evaluate values of $0.2$ and $2.0$ and also explore sampling $\lambda$ from the uniform distribution. In our investigation, $5$ and $10$ emerge as well-performing choices for the majority of algorithms when considering the $k$ parameter. Regarding $\boldsymbol{\sigma}$, the most suitable choice varies a lot among different algorithms, indicating distinct levels of robustness under varying noise scales. For $\alpha$, $0.2$ emerges as a suitable choice for most algorithms. Overall, NNG-Mix demonstrates robustness across different parameter settings, displaying fluctuations of less than $2\%$ in most cases.

\subsection{Visualization of the Generated Pseudo-anomalies}

\begin{figure*}[t!]
  \centering  \includegraphics[width=0.7\linewidth]{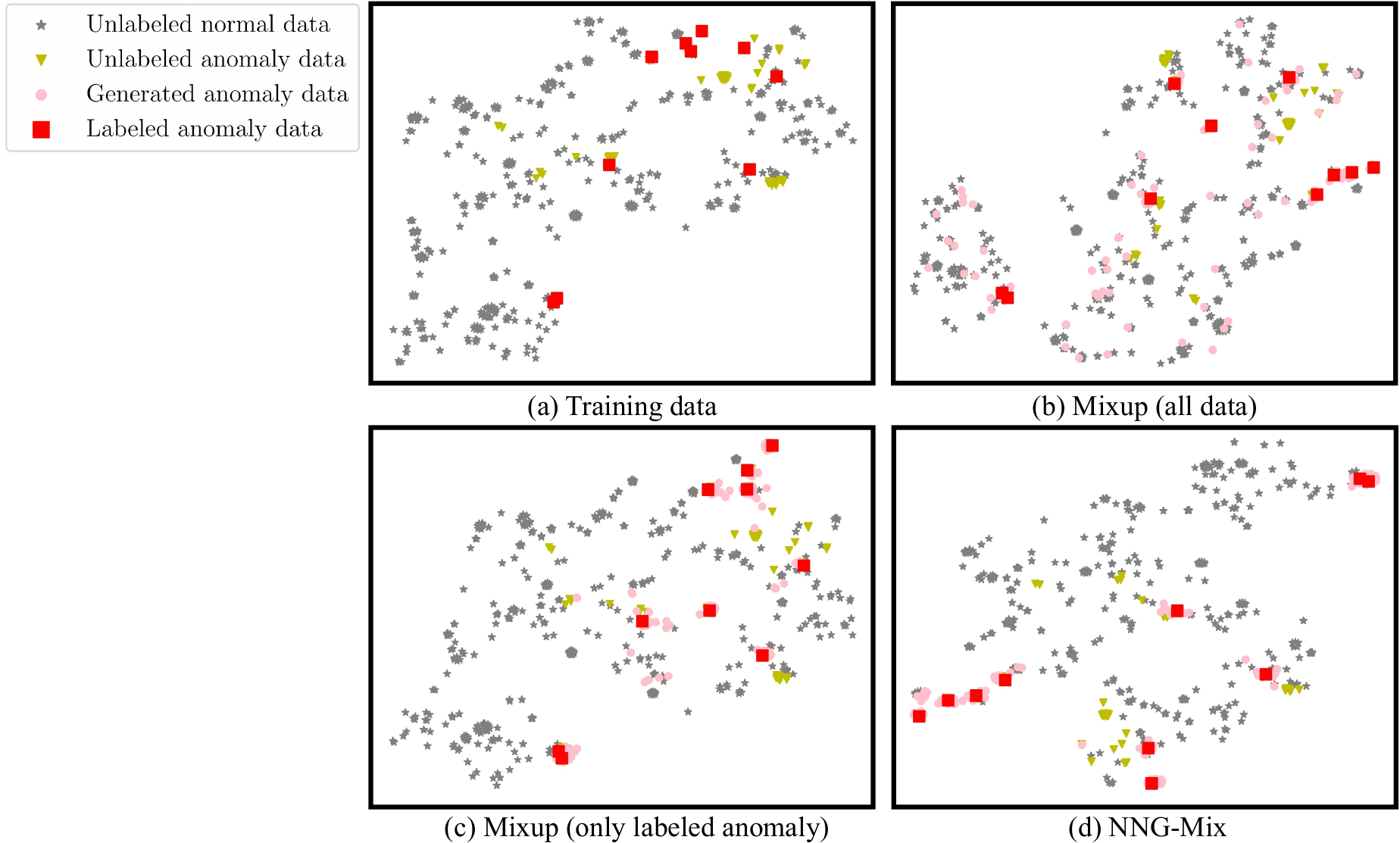}
   \caption{t-SNE visualization of pseudo-anomaly generation using Mixup~\cite{zhang2018mixup} and the proposed NNG-Mix on the vertebral dataset within the Classical dataset. Mixup often introduces unwanted noise samples within the distribution of unlabeled data, potentially significantly hindering the performance of AD models. In contrast, our NNG-Mix excels at generating pseudo-anomalies by effectively utilizing information from both labeled and unlabeled data.
   }
   \label{fig:vis_gen_crop}
\end{figure*}

Results in \cref{fig:vis_gen_crop} showcase the t-SNE embedding \cite{van2008visualizing} of pseudo-anomaly generation using both Mixup and our proposed NNG-Mix on the vertebral dataset within the Classical dataset. The unlabeled data includes both unlabeled normal data and unlabeled anomaly data, which are all treated as normal data during pseudo-anomaly generation and training. We use Mixup and NNG-Mix to generate more anomaly data using both labeled anomalies and unlabeled data. Application of Mixup across all training data introduces unwanted noise samples during the training process. Employing Mixup solely with labeled anomalies tends to overlook valuable information from unlabeled data and also introduces samples surrounded by unlabeled data, which are unsuitable for considering these as anomalies. In contrast, NNG-Mix adeptly harnesses information from both labeled anomalies and unlabeled data to effectively generate pseudo-anomalies.

\begin{figure}[t!]
  \centering  \includegraphics[width=\linewidth]{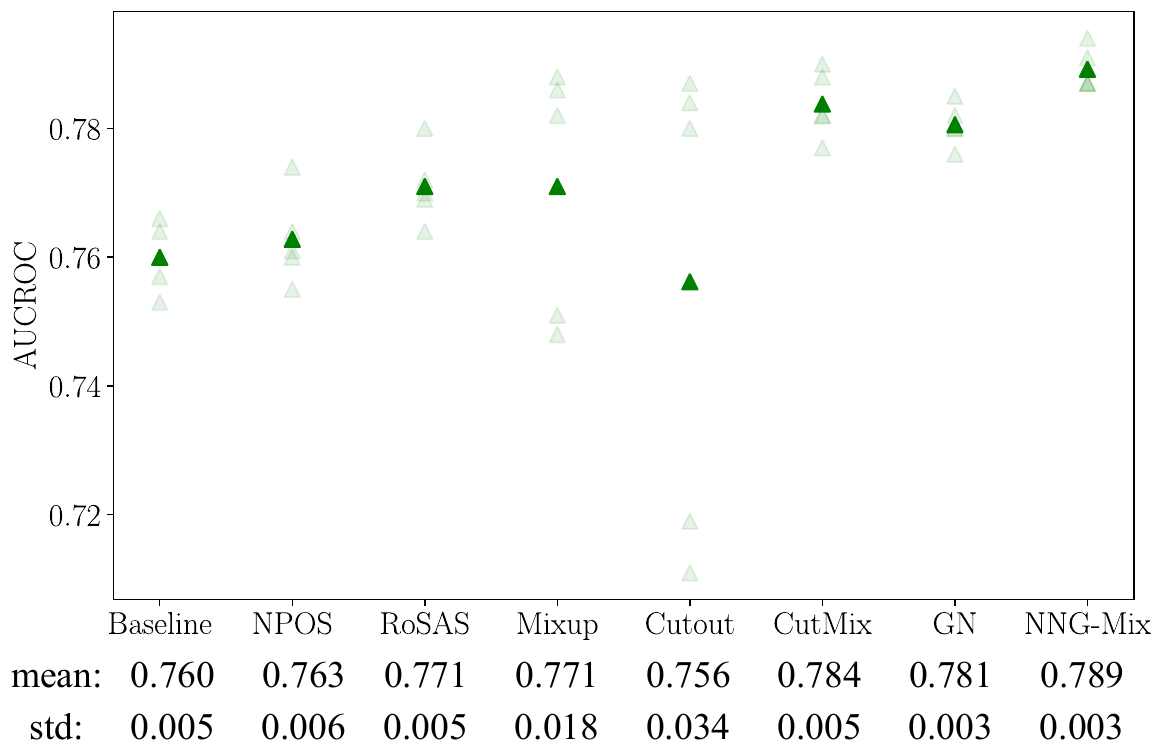}
   \caption{Experiments conducted with five random seeds on Classical datasets with $1\%$ available labeled anomalies and $10 \times$ pseudo-anomaly generation using DeepSAD. Bold foreground points  represent  average results across the five  seeds, while faint  background points depict  results from each   individual seed. NNG-Mix consistently  outperforms comparison methods  across different random seeds.}
   \label{fig:seeds}
\end{figure}

\subsection{Experiments with Different Random Seeds}
We conducted each experiment five times using different random seeds on Classical datasets with $1\%$ available labeled anomalies and $10 \times$ pseudo-anomaly generation using DeepSAD. We then calculate the mean and standard deviation of AUCROC to ensure that the results are robust and not due to chances.
As shown in~\cref{fig:seeds}, NNG-Mix exhibits consistency and significantly outperforms the baselines across different random seeds.

\begin{figure*}[t!]
  \centering  \includegraphics[width=\linewidth]{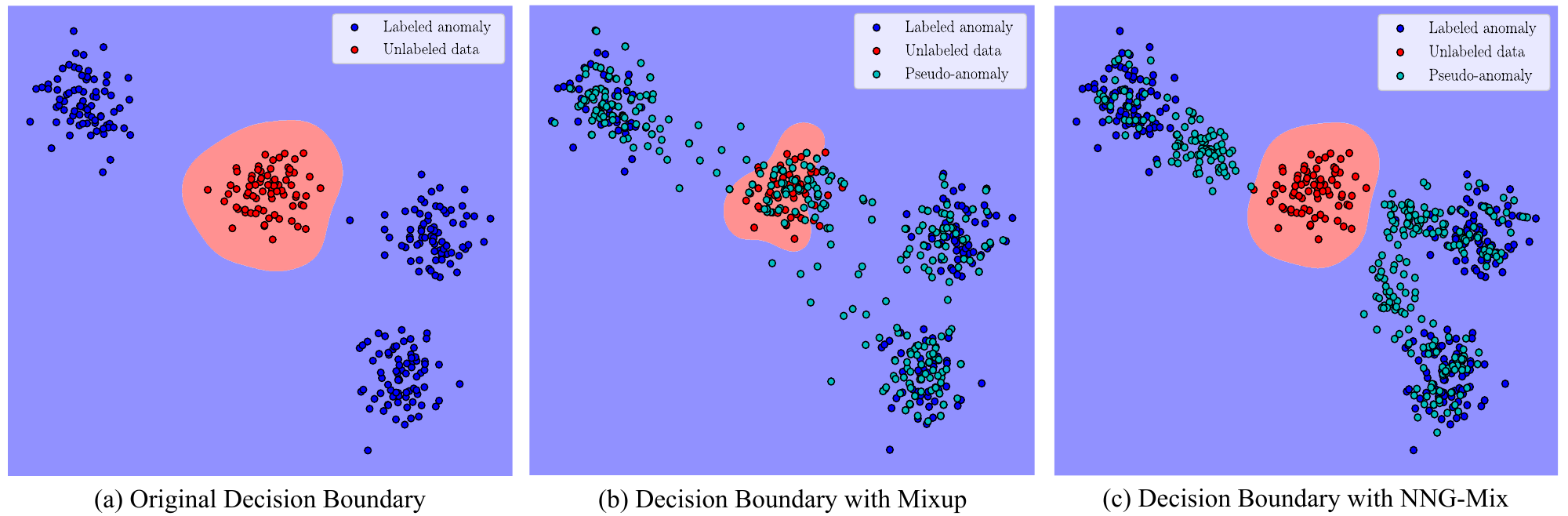}
   \caption{Influence of pseudo-anomalies on classifier decision boundaries. Mixup can cause  manifold intrusion by generating pseudo-anomalies within the distribution of unlabeled data, distorting  the original decision boundary. In contrast,  pseudo-anomalies generated by NNG-Mix are positioned either within the distribution of originally labeled anomalies or in the intermediate space between labeled anomalies and unlabeled data. This placement allows the classifier to form  a more compact  and effective decision boundary, avoiding the negative impacts of manifold intrusion.  }
   \label{fig:db}
\end{figure*}

\subsection{In-depth Analysis and Discussion on NNG-Mix}
\subsubsection{Influence of Pseudo-anomalies on Classifier Decision Boundary} In this section, we  examine the potential  negative impact of generated pseudo-anomalies on the classifier's decision boundary. We consider a straightforward  scenario involving  three clusters of labeled anomalies and one cluster of unlabeled data. The unlabeled data are centrally located among the labeled anomalies, as depicted in~\cref{fig:db} (a). We label the anomalies with $y=1$ and the unlabeled data with $y=0$, and train a Support Vector Machine (SVM)~\cite{hearst1998support} on this dataset. \cref{fig:db} (a) illustrates the initial decision boundary formed by the SVM  using the original data. 

Subsequently, we generate pseudo-anomalies using both Mixup and our NNG-Mix, labeling these pseudo-anomalies with $y=1$. These pseudo-anomalies are then incorporated into the original dataset for training new SVM models. \cref{fig:db} (b) and \cref{fig:db} (c) display  the decision boundaries of the SVM with pseudo-anomalies generated by Mixup and NNG-Mix, respectively. 

Mixup, which randomly selects two samples from labeled anomalies for mixing, often introduces manifold intrusion by placing  pseudo-anomalies within the distribution of unlabeled data. This intrusion  distorts  the original decision boundary, leading to potential  misclassification  of unlabeled data, as shown in~\cref{fig:db} (b). In contrast, NNG-Mix  strategically generates pseudo-anomalies by  mixing a sample only with its top-$k$ nearest neighbors. This approach ensures that pseudo-anomalies are either nestled within the cluster of originally labeled anomalies or situated in the intermediate space between labeled anomalies and unlabeled data. Consequently, the SVM trained with NNG-Mix-generated data  learns a more  precise and compact decision boundary, effectively avoiding the  negative effects of manifold intrusion, as shown in~\cref{fig:db} (c). This indicates that leveraging nearest-neighbor samples for pseudo-anomaly generation is a more effective strategy, avoiding the manifold intrusion that could potentially impact decision boundaries. 


\subsubsection{Major Differences between NNG-Mix and Compared Methods}  NPOS~\cite{tao2023nonparametric} is a recent anomaly synthesis algorithm designed for out-of-distribution detection~\cite{dong2024multiood}. The fundamental  difference between NNG-Mix and NPOS lies in the use of labeled anomalies during the pseudo-anomaly generation process. NNG-Mix utilizes a few labeled anomalies, which allows it to effectively harness both labeled and unlabeled data. Specifically, it generates anomaly samples in the intermediate space between labeled anomalies and unlabeled data, thereby leveraging a broader feature space, as depicted in~\cref{fig:db} (c). In contrast, NPOS operates under the assumption that only normal samples are available and consequently generates pseudo-anomalies close to these normal samples. This approach restricts  NPOS from fully utilizing  the broader feature space to generate diverse pseudo-anomalies. Another significant  difference is the specific  use of nearest-neighbor samples. NNG-Mix exclusively  mixes a sample with its top-$k$ nearest neighbors to prevent  manifold intrusion, whereas NPOS uses nearest-neighbor distances as a heuristic to identify  boundary normal samples and subsequently  generates pseudo-anomalies around these boundary normal samples. 

Compared to Mixup, the major advantage of NNG-Mix lies in its  strategic use of nearest-neighbor information to prevent manifold intrusion. As shown in~\cref{fig:db} (b), Mixup introduces manifold intrusion by generating pseudo-anomalies within the distribution of unlabeled data, which distorts the original decision boundary. In contrast, NNG-Mix generates pseudo-anomalies in a manner that enables  the classifier  to establish a more compact decision boundary, effectively avoiding  the negative impacts  of manifold intrusion.

In summary,  NNG-Mix's ability to \textit{explore a broader feature space} with  limited labeled
anomalies, along  with its \textit{avoidance of manifold intrusion} through the use of nearest-neighbor information, makes NNG-Mix superior to other baselines in most cases.

\section{Conclusions}
In this paper, we investigate improving semi-supervised anomaly detection performances from a novel view and propose the NNG-Mix algorithm for effective pseudo-anomaly generation. Our approach maximizes the utilization of information from both labeled anomalies and unlabeled data, resulting in better pseudo-anomaly generation for training in comparison to other data augmentation methods. As a result, NNG-Mix improves the generalization ability of SOTA semi-supervised and supervised AD algorithms. We conducted extensive experiments on $57$ datasets in ADBench and evaluated five semi-supervised and supervised anomaly detection algorithms. Our study offers a comprehensive comparison of various data augmentation algorithms for generating anomalies and demonstrates the consistently superior performance of NNG-Mix compared to various baselines across all anomaly detection algorithms. Our work presents a novel perspective for enhancing the performance of both semi-supervised and supervised AD algorithms. In future work, it is also interesting to investigate pseudo-anomaly generation on graphs and videos.

{\small
\bibliographystyle{IEEEtran}
\bibliography{egbib}

\begin{thebibliography}{10}
\providecommand{\url}[1]{#1}
\csname url@samestyle\endcsname
\providecommand{\newblock}{\relax}
\providecommand{\bibinfo}[2]{#2}
\providecommand{\BIBentrySTDinterwordspacing}{\spaceskip=0pt\relax}
\providecommand{\BIBentryALTinterwordstretchfactor}{4}
\providecommand{\BIBentryALTinterwordspacing}{\spaceskip=\fontdimen2\font plus
\BIBentryALTinterwordstretchfactor\fontdimen3\font minus \fontdimen4\font\relax}
\providecommand{\BIBforeignlanguage}[2]{{%
\expandafter\ifx\csname l@#1\endcsname\relax
\typeout{** WARNING: IEEEtran.bst: No hyphenation pattern has been}%
\typeout{** loaded for the language `#1'. Using the pattern for}%
\typeout{** the default language instead.}%
\else
\language=\csname l@#1\endcsname
\fi
#2}}
\providecommand{\BIBdecl}{\relax}
\BIBdecl

\bibitem{yu2017ring}
W.~Yu, J.~Li, M.~Z.~A. Bhuiyan, R.~Zhang, and J.~Huai, ``Ring: Real-time emerging anomaly monitoring system over text streams,'' \emph{IEEE Transactions on Big Data}, vol.~5, no.~4, pp. 506--519, 2017.

\bibitem{zhao2021suod}
Y.~Zhao, X.~Hu, C.~Cheng, C.~Wang, C.~Wan, W.~Wang, J.~Yang, H.~Bai, Z.~Li, C.~Xiao \emph{et~al.}, ``Suod: Accelerating large-scale unsupervised heterogeneous outlier detection,'' \emph{Proceedings of Machine Learning and Systems}, vol.~3, pp. 463--478, 2021.

\bibitem{zhou2020siamese}
X.~Zhou, W.~Liang, S.~Shimizu, J.~Ma, and Q.~Jin, ``Siamese neural network based few-shot learning for anomaly detection in industrial cyber-physical systems,'' \emph{IEEE Transactions on Industrial Informatics}, vol.~17, no.~8, pp. 5790--5798, 2020.

\bibitem{frusque2023non}
G.~Frusque, D.~Mitchell, J.~Blanche, D.~Flynn, and O.~Fink, ``Non-contact sensing for anomaly detection in wind turbine blades: A focus-svdd with complex-valued auto-encoder approach,'' \emph{arXiv preprint arXiv:2306.10808}, 2023.

\bibitem{zhao2023dynamic}
M.~Zhao and O.~Fink, ``Dynamic graph attention for anomaly detection in heterogeneous sensor networks,'' \emph{arXiv preprint arXiv:2307.03761}, 2023.

\bibitem{michau2021unsupervised}
G.~Michau and O.~Fink, ``Unsupervised transfer learning for anomaly detection: Application to complementary operating condition transfer,'' \emph{Knowledge-Based Systems}, vol. 216, p. 106816, 2021.

\bibitem{bogdoll2022anomaly}
D.~Bogdoll, M.~Nitsche, and J.~M. Z{\"o}llner, ``Anomaly detection in autonomous driving: A survey,'' in \emph{Proceedings of the IEEE/CVF conference on computer vision and pattern recognition}, 2022, pp. 4488--4499.

\bibitem{SuperFusion}
H.~Dong, X.~Zhang, J.~Xu, R.~Ai, W.~Gu, H.~Lu, J.~Kannala, and X.~Chen, ``Superfusion: Multilevel lidar-camera fusion for long-range hd map generation,'' \emph{arXiv preprint arXiv:2211.15656}, 2022.

\bibitem{dong2023jras}
H.~Dong, X.~Chen, S.~S{\"a}rkk{\"a}, and C.~Stachniss, ``Online pole segmentation on range images for long-term lidar localization in urban environments,'' \emph{Robotics and Autonomous Systems}, vol. 159, p. 104283, 2023.

\bibitem{han2022adbench}
S.~Han, X.~Hu, H.~Huang, M.~Jiang, and Y.~Zhao, ``Adbench: Anomaly detection benchmark,'' in \emph{Neural Information Processing Systems (NeurIPS)}, 2022.

\bibitem{ruff2018deep}
L.~Ruff, R.~Vandermeulen, N.~Goernitz, L.~Deecke, S.~A. Siddiqui, A.~Binder, E.~M{\"u}ller, and M.~Kloft, ``Deep one-class classification,'' in \emph{International conference on machine learning}.\hskip 1em plus 0.5em minus 0.4em\relax PMLR, 2018, pp. 4393--4402.

\bibitem{ergen2019unsupervised}
T.~Ergen and S.~S. Kozat, ``Unsupervised anomaly detection with lstm neural networks,'' \emph{IEEE transactions on neural networks and learning systems}, vol.~31, no.~8, pp. 3127--3141, 2019.

\bibitem{liu2021anomaly}
Y.~Liu, Z.~Li, S.~Pan, C.~Gong, C.~Zhou, and G.~Karypis, ``Anomaly detection on attributed networks via contrastive self-supervised learning,'' \emph{IEEE transactions on neural networks and learning systems}, vol.~33, no.~6, pp. 2378--2392, 2021.

\bibitem{scholkopf2001estimating}
B.~Sch{\"o}lkopf, J.~C. Platt, J.~Shawe-Taylor, A.~J. Smola, and R.~C. Williamson, ``Estimating the support of a high-dimensional distribution,'' \emph{Neural computation}, vol.~13, no.~7, pp. 1443--1471, 2001.

\bibitem{vandermeulen2013consistency}
R.~Vandermeulen and C.~Scott, ``Consistency of robust kernel density estimators,'' in \emph{Conference on Learning Theory}.\hskip 1em plus 0.5em minus 0.4em\relax PMLR, 2013, pp. 568--591.

\bibitem{liu2008isolation}
F.~T. Liu, K.~M. Ting, and Z.-H. Zhou, ``Isolation forest,'' in \emph{2008 eighth ieee international conference on data mining}.\hskip 1em plus 0.5em minus 0.4em\relax IEEE, 2008, pp. 413--422.

\bibitem{li2022ecod}
Z.~Li, Y.~Zhao, X.~Hu, N.~Botta, C.~Ionescu, and G.~Chen, ``Ecod: Unsupervised outlier detection using empirical cumulative distribution functions,'' \emph{IEEE Transactions on Knowledge and Data Engineering}, 2022.

\bibitem{zong2018deep}
B.~Zong, Q.~Song, M.~R. Min, W.~Cheng, C.~Lumezanu, D.~Cho, and H.~Chen, ``Deep autoencoding gaussian mixture model for unsupervised anomaly detection,'' in \emph{International conference on learning representations}, 2018.

\bibitem{ramaswamy2000efficient}
S.~Ramaswamy, R.~Rastogi, and K.~Shim, ``Efficient algorithms for mining outliers from large data sets,'' in \emph{Proceedings of the 2000 ACM SIGMOD international conference on Management of data}, 2000, pp. 427--438.

\bibitem{breunig2000lof}
M.~M. Breunig, H.-P. Kriegel, R.~T. Ng, and J.~Sander, ``Lof: identifying density-based local outliers,'' in \emph{Proceedings of the 2000 ACM SIGMOD international conference on Management of data}, 2000, pp. 93--104.

\bibitem{DeepSAD}
L.~Ruff, R.~A. Vandermeulen, N.~G{\"{o}}rnitz, A.~Binder, E.~M{\"{u}}ller, K.~M{\"{u}}ller, and M.~Kloft, ``Deep semi-supervised anomaly detection,'' in \emph{International conference on learning representations}, 2020.

\bibitem{pang2023deep}
G.~Pang, C.~Shen, H.~Jin, and A.~van~den Hengel, ``Deep weakly-supervised anomaly detection,'' in \emph{Proceedings of the 29th ACM SIGKDD Conference on Knowledge Discovery and Data Mining}, 2023, pp. 1795--1807.

\bibitem{pang2019deep}
G.~Pang, C.~Shen, and A.~Van Den~Hengel, ``Deep anomaly detection with deviation networks,'' in \emph{Proceedings of the 25th ACM SIGKDD international conference on knowledge discovery \& data mining}, 2019, pp. 353--362.

\bibitem{mpad}
S.~Zhao, Z.~Yu, S.~Li, X.~Wang, T.~G. Marbach, G.~Wang, and X.~Liu, ``Meta pseudo labels for anomaly detection via partially observed anomalies,'' \emph{Engineering Applications of Artificial Intelligence}, vol. 126, 2024.

\bibitem{zhou2021feature}
Y.~Zhou, X.~Song, Y.~Zhang, F.~Liu, C.~Zhu, and L.~Liu, ``Feature encoding with autoencoders for weakly supervised anomaly detection,'' \emph{IEEE Transactions on Neural Networks and Learning Systems}, vol.~33, no.~6, pp. 2454--2465, 2021.

\bibitem{zhao2018xgbod}
Y.~Zhao and M.~K. Hryniewicki, ``Xgbod: improving supervised outlier detection with unsupervised representation learning,'' in \emph{2018 International Joint Conference on Neural Networks (IJCNN)}.\hskip 1em plus 0.5em minus 0.4em\relax IEEE, 2018, pp. 1--8.

\bibitem{rosenblatt1958perceptron}
F.~Rosenblatt, ``The perceptron: a probabilistic model for information storage and organization in the brain.'' \emph{Psychological review}, vol.~65, no.~6, p. 386, 1958.

\bibitem{gorishniy2021revisiting}
Y.~Gorishniy, I.~Rubachev, V.~Khrulkov, and A.~Babenko, ``Revisiting deep learning models for tabular data,'' in \emph{Neural Information Processing Systems (NeurIPS)}, 2021.

\bibitem{prokhorenkova2018catboost}
L.~Prokhorenkova, G.~Gusev, A.~Vorobev, A.~V. Dorogush, and A.~Gulin, ``Catboost: unbiased boosting with categorical features,'' in \emph{Neural Information Processing Systems (NeurIPS)}, 2018.

\bibitem{aggarwal2017introduction}
C.~C. Aggarwal and C.~C. Aggarwal, \emph{An introduction to outlier analysis}.\hskip 1em plus 0.5em minus 0.4em\relax Springer, 2017.

\bibitem{massoli2021mocca}
F.~V. Massoli, F.~Falchi, A.~Kantarci, {\c{S}}.~Akti, H.~K. Ekenel, and G.~Amato, ``Mocca: Multilayer one-class classification for anomaly detection,'' \emph{IEEE Transactions on Neural Networks and Learning Systems}, vol.~33, no.~6, pp. 2313--2323, 2021.

\bibitem{zhang2018mixup}
H.~Zhang, M.~Cisse, Y.~N. Dauphin, and D.~Lopez-Paz, ``mixup: Beyond empirical risk minimization,'' in \emph{International conference on learning representations}, 2018.

\bibitem{devries2017cutout}
T.~DeVries and G.~W. Taylor, ``Improved regularization of convolutional neural networks with cutout,'' \emph{arXiv preprint arXiv:1708.04552}, 2017.

\bibitem{yun2019cutmix}
S.~Yun, D.~Han, S.~J. Oh, S.~Chun, J.~Choe, and Y.~Yoo, ``Cutmix: Regularization strategy to train strong classifiers with localizable features,'' in \emph{Proceedings of the IEEE/CVF conference on computer vision and pattern recognition}, 2019.

\bibitem{bergman2019classification}
L.~Bergman and Y.~Hoshen, ``Classification-based anomaly detection for general data,'' in \emph{International conference on learning representations}, 2020.

\bibitem{zhao2019pyod}
Y.~Zhao, Z.~Nasrullah, and Z.~Li, ``Pyod: A python toolbox for scalable outlier detection,'' \emph{JMLR}, vol.~20, pp. 1--7, 2019.

\bibitem{vaswani2017attention}
A.~Vaswani, N.~Shazeer, N.~Parmar, J.~Uszkoreit, L.~Jones, A.~N. Gomez, {\L}.~Kaiser, and I.~Polosukhin, ``Attention is all you need,'' \emph{Advances in neural information processing systems}, vol.~30, 2017.

\bibitem{xu2023rosas}
H.~Xu, Y.~Wang, G.~Pang, S.~Jian, N.~Liu, and Y.~Wang, ``Rosas: Deep semi-supervised anomaly detection with contamination-resilient continuous supervision,'' \emph{Information Processing \& Management}, vol.~60, no.~5, p. 103459, 2023.

\bibitem{yang2022image}
S.~Yang, W.~Xiao, M.~Zhang, S.~Guo, J.~Zhao, and F.~Shen, ``Image data augmentation for deep learning: A survey,'' \emph{arXiv preprint arXiv:2204.08610}, 2022.

\bibitem{feng2021survey}
S.~Y. Feng, V.~Gangal, J.~Wei, S.~Chandar, S.~Vosoughi, T.~Mitamura, and E.~Hovy, ``A survey of data augmentation approaches for nlp,'' \emph{arXiv preprint arXiv:2105.03075}, 2021.

\bibitem{chen2020simple}
T.~Chen, S.~Kornblith, M.~Norouzi, and G.~Hinton, ``A simple framework for contrastive learning of visual representations,'' in \emph{International conference on machine learning}.\hskip 1em plus 0.5em minus 0.4em\relax PMLR, 2020, pp. 1597--1607.

\bibitem{yao2022cmix}
H.~Yao, Y.~Wang, L.~Zhang, J.~Zou, and C.~Finn, ``C-mixup: Improving generalization in regression,'' in \emph{Proceeding of the Thirty-Sixth Conference on Neural Information Processing Systems}, 2022.

\bibitem{li2022who}
C.~Li, X.~Li, L.~Feng, and J.~Ouyang, ``Who is your right mixup partner in positive and unlabeled learning,'' in \emph{International Conference on Learning Representations}, 2022.

\bibitem{cubuk2019autoaugment}
E.~D. Cubuk, B.~Zoph, D.~Mane, V.~Vasudevan, and Q.~V. Le, ``Autoaugment: Learning augmentation strategies from data,'' in \emph{Proceedings of the IEEE/CVF conference on computer vision and pattern recognition}, 2019, pp. 113--123.

\bibitem{cai2022perturbation}
J.~Cai and J.~Fan, ``Perturbation learning based anomaly detection,'' \emph{Advances in Neural Information Processing Systems}, vol.~35, pp. 14\,317--14\,330, 2022.

\bibitem{wei2019eda}
J.~Wei and K.~Zou, ``Eda: Easy data augmentation techniques for boosting performance on text classification tasks,'' \emph{arXiv preprint arXiv:1901.11196}, 2019.

\bibitem{guo2020nonlinear}
H.~Guo, ``Nonlinear mixup: Out-of-manifold data augmentation for text classification,'' in \emph{Proceedings of the AAAI Conference on Artificial Intelligence}, 2020.

\bibitem{xu2024calibrated}
H.~Xu, Y.~Wang, S.~Jian, Q.~Liao, Y.~Wang, and G.~Pang, ``Calibrated one-class classification for unsupervised time series anomaly detection,'' \emph{IEEE Transactions on Knowledge and Data Engineering}, 2024.

\bibitem{bergman2020classification}
L.~Bergman and Y.~Hoshen, ``Classification-based anomaly detection for general data,'' \emph{arXiv preprint arXiv:2005.02359}, 2020.

\bibitem{jing2020self}
L.~Jing and Y.~Tian, ``Self-supervised visual feature learning with deep neural networks: A survey,'' \emph{IEEE transactions on pattern analysis and machine intelligence}, vol.~43, no.~11, pp. 4037--4058, 2020.

\bibitem{li2021cutpaste}
C.-L. Li, K.~Sohn, J.~Yoon, and T.~Pfister, ``Cutpaste: Self-supervised learning for anomaly detection and localization,'' in \emph{Proceedings of the IEEE/CVF conference on computer vision and pattern recognition}, 2021, pp. 9664--9674.

\bibitem{hendrycks2019using}
D.~Hendrycks, M.~Mazeika, S.~Kadavath, and D.~Song, ``Using self-supervised learning can improve model robustness and uncertainty,'' \emph{Advances in neural information processing systems}, vol.~32, 2019.

\bibitem{sehwag2021ssd}
V.~Sehwag, M.~Chiang, and P.~Mittal, ``Ssd: A unified framework for self-supervised outlier detection,'' \emph{arXiv preprint arXiv:2103.12051}, 2021.

\bibitem{winkens2020contrastive}
J.~Winkens, R.~Bunel, A.~G. Roy, R.~Stanforth, V.~Natarajan, J.~R. Ledsam, P.~MacWilliams, P.~Kohli, A.~Karthikesalingam, S.~Kohl \emph{et~al.}, ``Contrastive training for improved out-of-distribution detection,'' \emph{arXiv preprint arXiv:2007.05566}, 2020.

\bibitem{tao2023nonparametric}
L.~Tao, X.~Du, J.~Zhu, and Y.~Li, ``Non-parametric outlier synthesis,'' in \emph{The Eleventh International Conference on Learning Representations}, 2023.

\bibitem{devlin2018bert}
J.~Devlin, M.-W. Chang, K.~Lee, and K.~Toutanova, ``Bert: Pre-training of deep bidirectional transformers for language understanding,'' \emph{arXiv preprint arXiv:1810.04805}, 2018.

\bibitem{wei2021open}
H.~Wei, L.~Tao, R.~Xie, and B.~An, ``Open-set label noise can improve robustness against inherent label noise,'' \emph{Advances in Neural Information Processing Systems}, vol.~34, pp. 7978--7992, 2021.

\bibitem{van2008visualizing}
L.~Van~der Maaten and G.~Hinton, ``Visualizing data using t-sne.'' \emph{Journal of machine learning research}, vol.~9, no.~11, 2008.

\bibitem{hearst1998support}
M.~A. Hearst, S.~T. Dumais, E.~Osuna, J.~Platt, and B.~Scholkopf, ``Support vector machines,'' \emph{IEEE Intelligent Systems and their applications}, vol.~13, no.~4, pp. 18--28, 1998.

\bibitem{dong2024multiood}
H.~Dong, Y.~Zhao, E.~Chatzi, and O.~Fink, ``{MultiOOD: Scaling Out-of-Distribution Detection for Multiple Modalities},'' \emph{arXiv preprint arXiv:2405.17419}, 2024.

\end{thebibliography}
}

\vfill

\end{document}